\documentclass[lettersize,journal]{IEEEtran}
\usepackage{amsmath,amsfonts}
\usepackage{algorithmic}
\usepackage{algorithm}
\usepackage{array}
\usepackage[caption=false,font=normalsize,labelfont=sf,textfont=sf]{subfig}
\usepackage{textcomp}
\usepackage{stfloats}
\usepackage{url}
\usepackage{verbatim}
\usepackage{graphicx}
\usepackage{cite}
\usepackage{xcolor}
\usepackage{soul}

\begin{document}

\title{BSL: Navigation Method Considering Blind Spots Based on ROS Navigation Stack and Blind Spots Layer for Mobile Robot
}
\author{Masato Kobayashi and Naoki Motoi
\thanks{
Masato Kobayashi is with the Cybermedia Center,
Osaka University, Japan.
Naoki Motoi is with the Graduate School of Maritime Sciences,
Kobe University, Japan.
}
}


\maketitle

\begin{abstract}
  This paper proposes a navigation method considering blind spots based on the robot operating system (ROS) navigation stack and blind spots layer (BSL) for a wheeled mobile robot.
  In this paper, environmental information is recognized using a laser range finder (LRF) and RGB-D cameras.
  Blind spots occur when corners or obstacles are present in the environment, and may lead to collisions if a human or object moves toward the robot from these blind spots.
  To prevent such collisions, this paper proposes a navigation method considering blind spots based on the local cost map layer of the BSL for the wheeled mobile robot.
  Blind spots are estimated by utilizing environmental data collected through RGB-D cameras.
  The navigation method that takes these blind spots into account is achieved through the implementation of the BSL and a local path planning method that employs an enhanced cost function of dynamic window approach.
  The effectiveness of the proposed method was further demonstrated through simulations and experiments.
\end{abstract}


\section{Introduction}
 \IEEEPARstart{A}{s} the application of autonomous mobile robots continues to proliferate, ensuring the coexistence of humans and robots is progressively becoming a central issue across a broad range of industries\cite{kobayashi2022bsl}.
 These robots find utility in various sectors, encompassing medical applications\mbox{\cite{Teeneti2021wheelchairs, wang2022medicalrobot}}, industrial settings \mbox{\cite{liao2023mwh, kumar2021surveyhrc}}, disaster \mbox{\cite{han2022srg, seeja2022snake}}, and food production \mbox{\cite{abegaz2022food, saito2021robot}}. It is noteworthy that the functionality of these robots is predominantly composed of two essential elements: mobility and manipulation\mbox{\cite{nagpal2018mani, aziz2019mani, martin2021mm1}}. Within the service sector, the necessity of ensuring secure and efficient interaction between humans and robots underlines the importance of judicious management of these elements\mbox {\cite{selvaggio2021hri}}.
 While manipulation remains a crucial facet, in this manuscript, we principally concentrate on the mobility aspect of the robots. Thus, the endeavor to address the challenge of human-robot coexistence, with a primary focus on robotic mobility, is essential in advancing the development of service robots\mbox{\cite{zhang2023service}}.

\begin{figure}[t]
  \begin{center}
    \scalebox{0.17}{
        \includegraphics{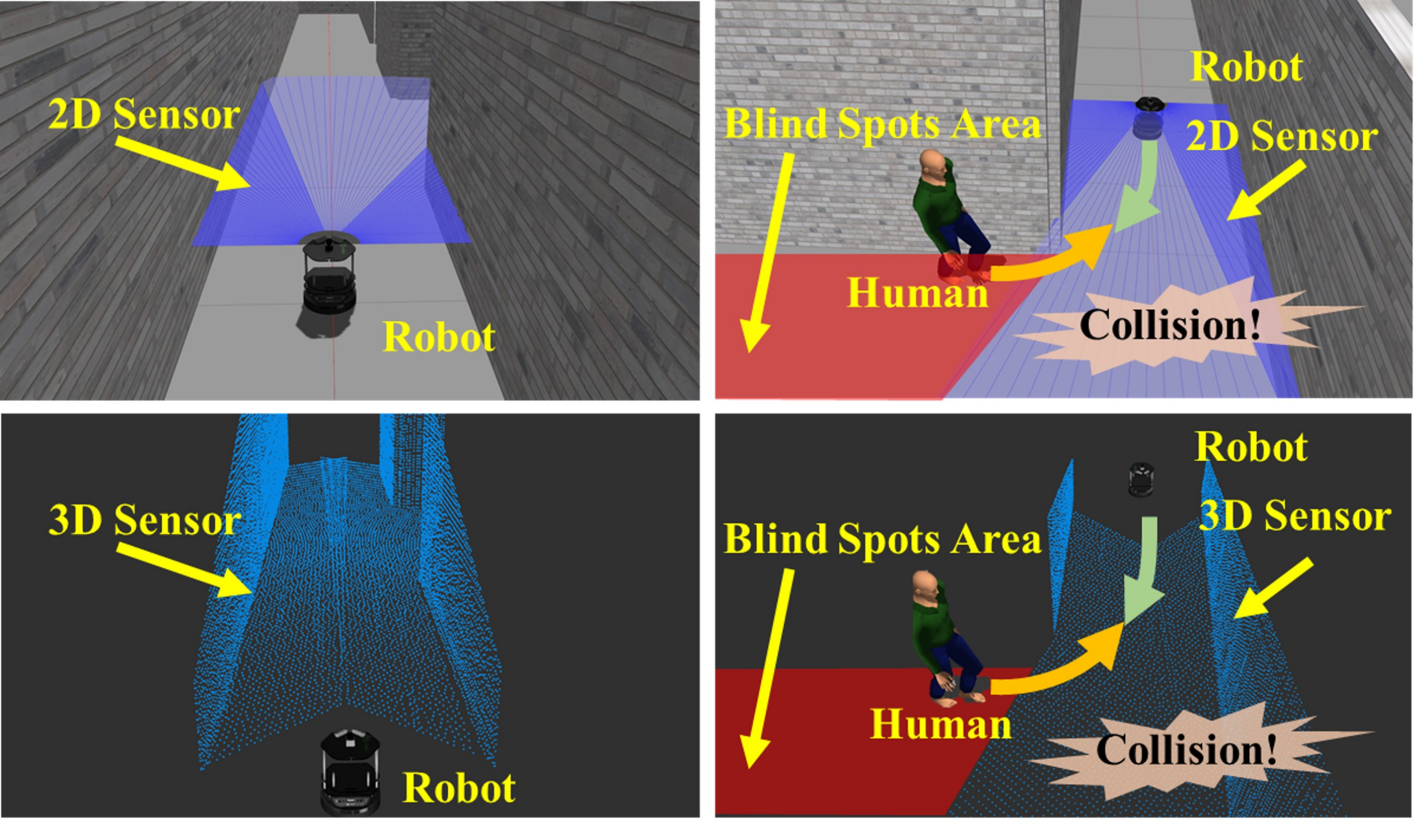}}    
\end{center}
  \vspace{-5mm}
  \caption{Image of Blind Spots Area} \label{fig:overview_bsl_rgbd}
\end{figure}

The typical configuration of an autonomous mobile robot system includes localization\cite{bae2022local}, mapping \cite{TIA-mapping}, perception \cite{TIA-perception}, and path planning \cite{TIA-pathplanning}.
To realize the coexistence of humans and robots in inhabited environments, it is imperative to generate paths for the robots that are devoid of collisions and adverse interactions with humans \cite{park2014robot, kurita2011motion,kobayashi2022dwv, mondal2020con}.
This paper focuses on the situations in which blind spots occur as the possibility of harming humans.
When there are obstacles in front of the robot or just before approaching the turn, blind spots are generated.
As shown in Fig.~\ref{fig:overview_bsl_rgbd}, when the human comes toward the robot from these blind spots, there is a high possibility that the robot will collide with the human\cite{schlegel2021bsl,zhu2020bsl,orzechowski2018bsl,hu2021medical}.

In conventional approaches for handling blind spots, real-time velocity control of the robot that accounts for these blind spots has been proposed \cite{1,2,3}. Furthermore, there are also path planning techniques that rely on maps to address blind spots \cite{4,5,6}. Despite these conventional methods, there are some challenges. Firstly, in many of these methods, the robot is only able to move along the pre-planned path, making it incapable of avoiding obstacles that are not present on the map. Secondly, these methods do not factor in collision avoidance and the constraints on the robot's motion.
In other words, a more flexible path planning method that detects blind spots, avoids obstacles, and takes into account the motion constraints of the robot in real-time is needed.
We proposed a local path planning method that addresses these needs, including blind spot detection, collision avoidance, and the robot's motion capabilities \cite{7}.
This system is based on the Navigation Stack of the Robot Operating System (ROS). The method employs a laser range finder (LRF) for blind spot detection, but the detection scope is restricted to the horizontal plane of the LRF, making it inflexible for a variety of environments. Thus, the ability to handle 3 dimensional (3D) information is required.

Many sensors such as RGB-D cameras and LiDAR are being used in mobile robots to acquire 3D environmental information.
RGB-D cameras provide both color (RGB) and depth (D) data. This dual-modality allows for detailed environmental mapping, object recognition, and pose estimation. Their relatively low cost and compact size make them ideal for the service robot application\mbox{\cite{rgbd-1,rgbd-2,rgbd-3}}. 
Furthermore, RGB-D cameras can effectively function in indoor environments, which is particularly beneficial for our study.
By providing 3D information, RGB-D cameras overcome the limitations of the LRF's horizontal detection scope.
As for the possibility of using other types of sensors, such as LiDAR, we acknowledge that LiDAR can offer more precise distance measurements and can function effectively in a variety of environments, including outdoors\mbox{\cite{LD-1,LD-2,LD-3}}.
However, LiDAR systems are typically more expensive and larger than RGB-D cameras or LRFs, which might be limiting factors for some applications.
In this paper, we used RGB-D cameras for getting 3D environment information.

This paper proposes the local path planning method based on the cost map by using RGB-D cameras\cite{kobayashi2022bsl}.
Our system is built upon the Robot Operating System (ROS) Navigation Stack.
The acquired point cloud data from RGB-D cameras are utilized to calculate the cost of blind spots, enabling real-time path planning that considers both the presence of blind spots and the motion constraints of the robot\cite{kobayashi2022bsl, dwa}.
This paper presents the effectiveness of the proposed method by introducing practice simulation environments where blind spots occur on both sides and experiments in the real world, which were not considered in the previous paper\cite{kobayashi2022bsl}.

The main contributions of our work are as follows.
 \begin{itemize}
  \item Our method introduces BSL, which dynamically estimates blind spot areas from 3D point cloud data, to achieve navigation that takes blind spot areas into account. 
  \item Our method is to add the blind spot area and robot velocity to the DWA evaluation function.
  \item Our method successfully considers blind spot area and robot constraint in both simulated and real-world experiments.
 \end{itemize}

This paper consists of eight sections including this one.
Section II shows the coordinate system.
Section III shows the navigation system.
Section IV explains the blind spots layer by LRF as the conventional method.
Section V proposes the blind spots layer by RGB-D cameras.
In Section VI, simulation results are shown to confirm the usefulness of the proposed method.
In Section VII, experiment results are shown to confirm the usefulness of the proposed method.
Section VIII concludes this paper.

\section{Coordinate System}
\begin{figure}[t]
  \begin{minipage}{1\hsize}
    \begin{center}
      \scalebox{0.23}{
        \includegraphics{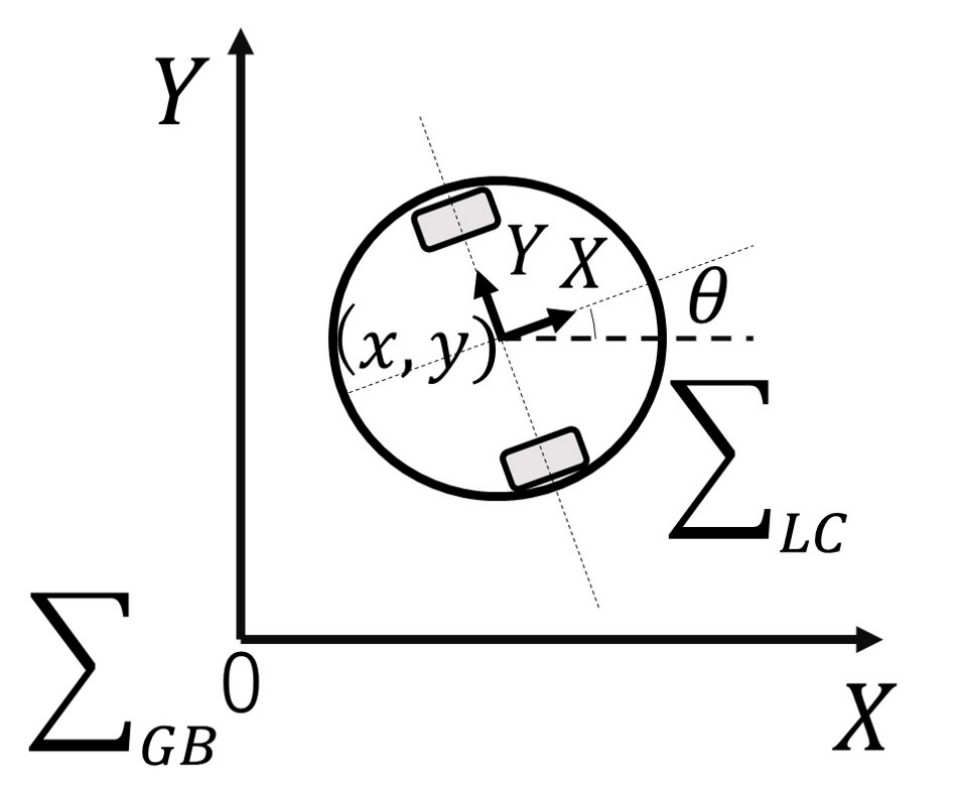}}
    \end{center}
  \end{minipage}
  \caption{Coordinate System}
  \label{fig:csa}
  \vspace{-4mm}
\end{figure}

Fig.~\ref{fig:csa} shows the coordinate system of the robot.
This paper defines the local coordinate system $\Sigma_{LC}$ and the global coordinate system $\Sigma_{GB}$.
The value in the global coordinate system is expressed as the superscript $^{GB}\bigcirc$.
The variable of the local coordinate system does not have the superscript.
The origin of the global coordinate system is set as an initial robot position.
The origin of the local coordinate system is set as the center point of both wheels.

\section{Navigation System}
\subsection{ROS Navigation Stack}
\begin{figure}[t]
  \begin{minipage}{1\hsize}
    \begin{center}
      \scalebox{0.21}{
        \includegraphics{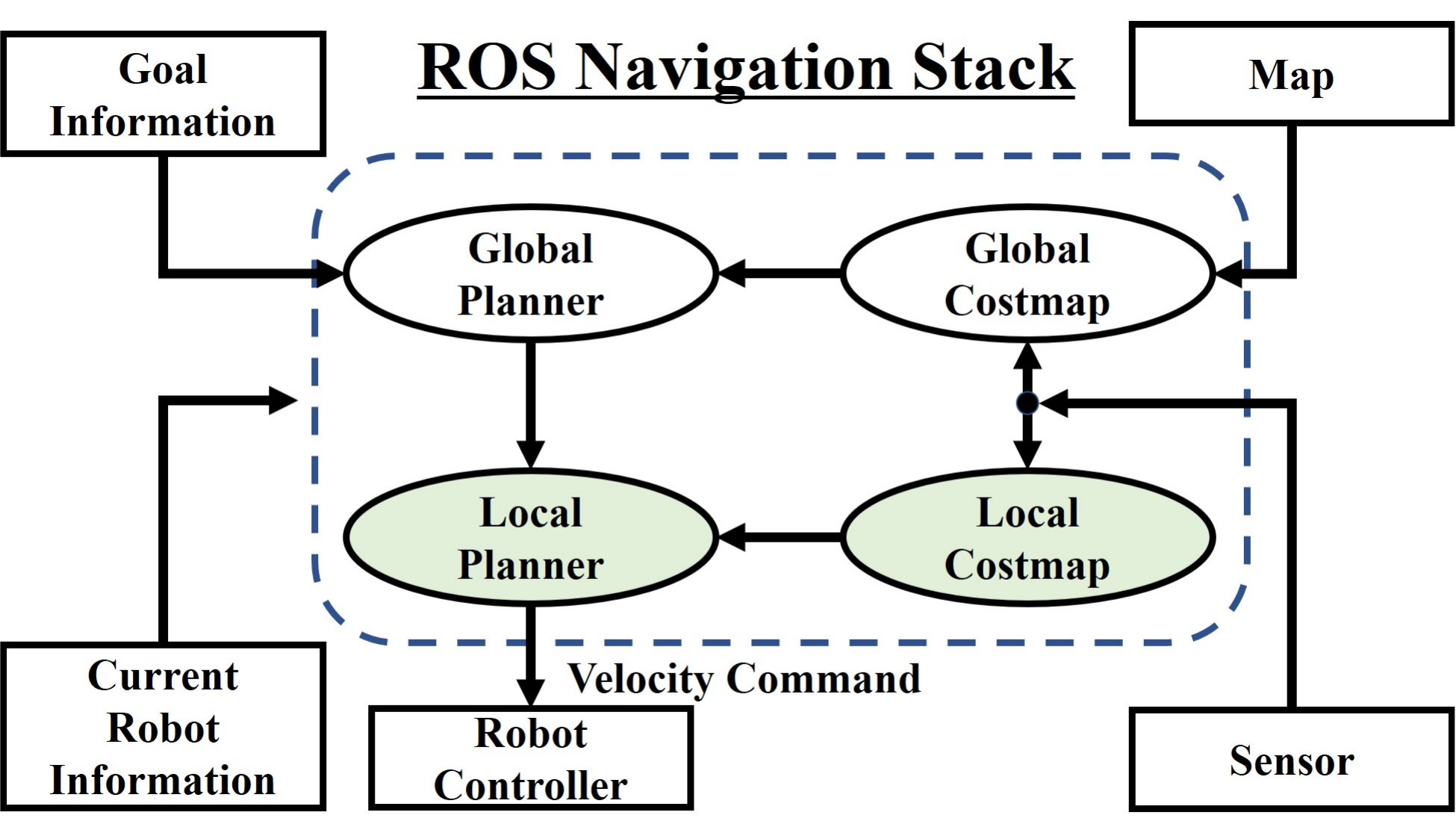}}
    \end{center}
  \end{minipage}
  \caption{ROS Navigation System}
  \label{fig:csb}
  \vspace{-4mm}
\end{figure}
ROS Navigation Stack is configured as shown in Fig.~\ref{fig:csb}.
The global cost map is calculated based on the map generated by the Simultaneous Localization and Mapping (SLAM).
Global path planning is performed to the destination by using the global cost map.
The local cost map is calculated from the information obtained from the sensors in real-time.
In order to avoid collisions with obstacles, the robot motion is determined by local path planning using the local cost map along the global path.
This paper focuses on the local path planning and the local cost map to achieve path planning that takes blind spots and robot motion constraints into account.
\subsection{Local Path Planning: DWA}
Dynamic window approach (DWA) calculates the Dynamic Window (DW), which is the range of possible motions determined by the specifications of the robot\cite{dwa}.
DWA calculates the position and posture after predicted time $T^{pre}$ by assuming constant translation and angular velocity within the DW.
The local path planning method adapts the calculated values to the cost function and selects the translation and angular velocity with the smallest cost function value.
\subsection{Cost Function}
The cost function used in the navigation stack is as follows.
\begin{equation}
  \begin{split}
    J=W^{pos}\cdot c^{pos}+W^{gol}\cdot c^{gol}+W^{obs}\cdot c^{obs}
  \end{split}
  \label{eq:j_def}
\end{equation}
where
$J$, $c^{pos}$, $c^{gol}$ and $c^{obs}$ represent for the total cost,
the distance from the local path endpoint to the global path,
the distance from the local path endpoint to the goal, and
the maximum map cost considering obstacles on the local path, respectively.
$W^{pos}$, $W^{gol}$ and $W^{obs}$ represent the weight coefficient for the global path,
the goal position, and the maximum obstacle cost on the local path, respectively.
\subsection{Local cost map}
\begin{figure}[t]
  \begin{minipage}{0.49\hsize}
    \begin{center}
      \scalebox{0.18}{
        \includegraphics{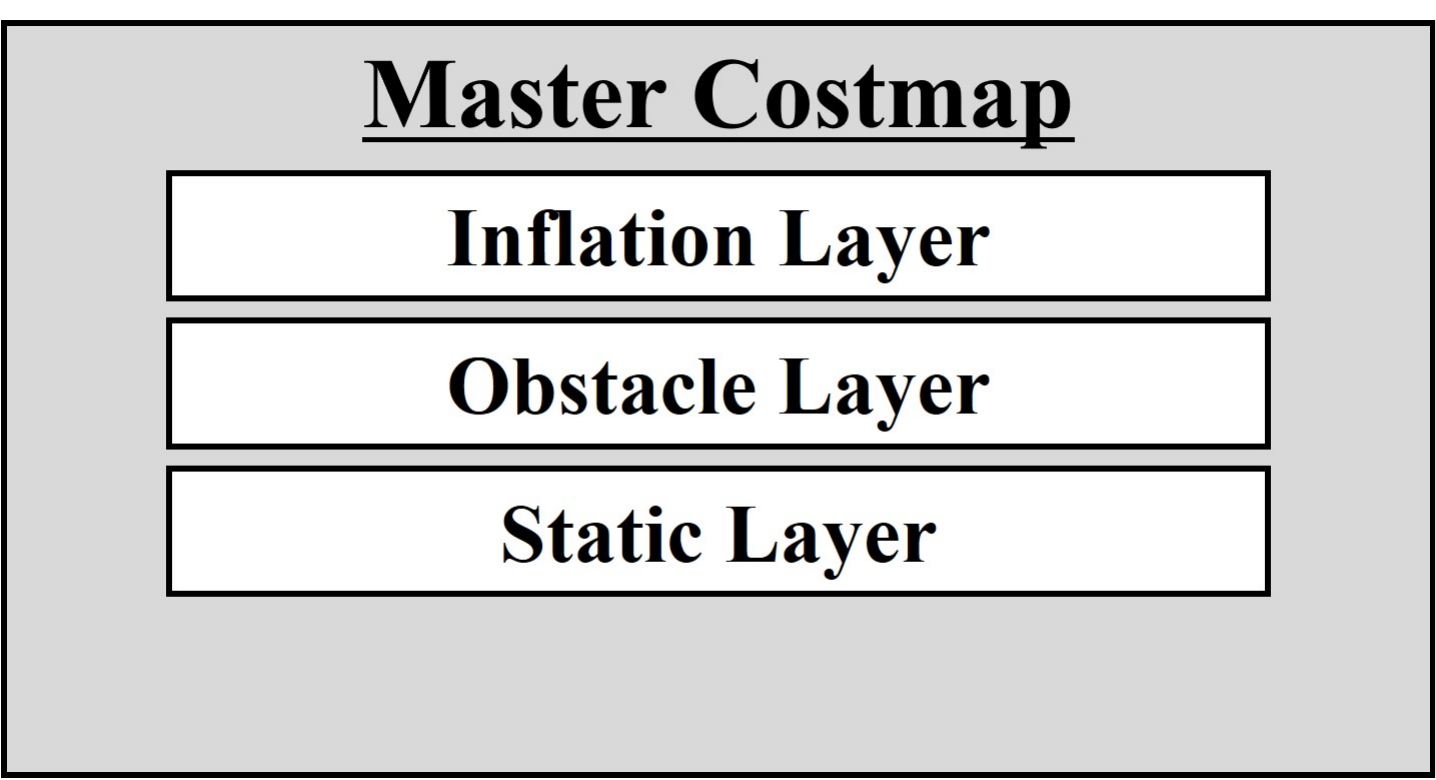}}
      {\begin{center} (a) Master cost map \end{center}}
    \end{center}
  \end{minipage}
  \begin{minipage}{0.49\hsize}
    \begin{center}
      \scalebox{0.16}{
        \includegraphics{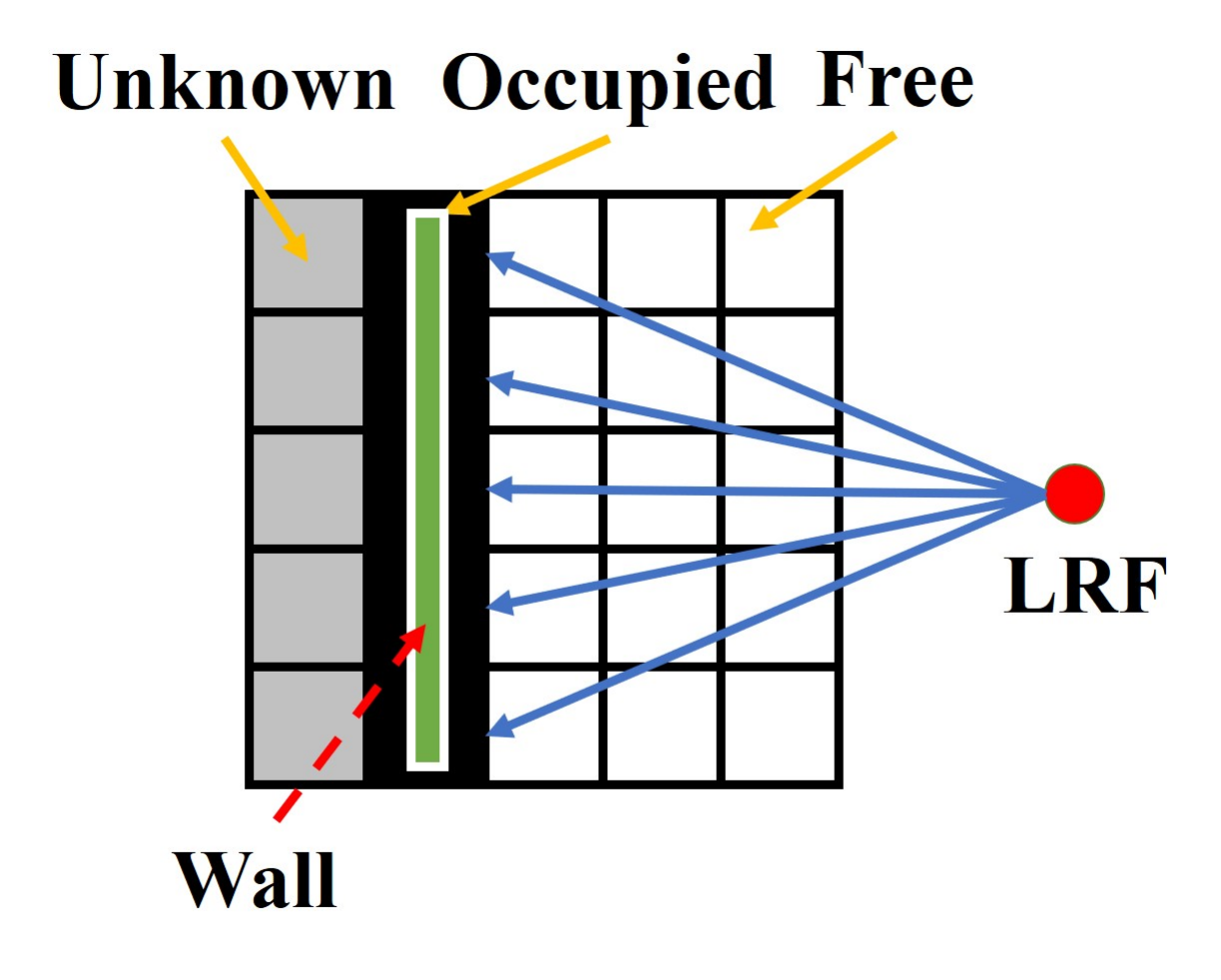}}
          \vspace{-4mm}
      {\begin{center} (b) Static Layer \end{center}}
    \end{center}
  \end{minipage}
  \begin{minipage}{0.49\hsize}
    \begin{center}
      \scalebox{0.16}{
        \includegraphics{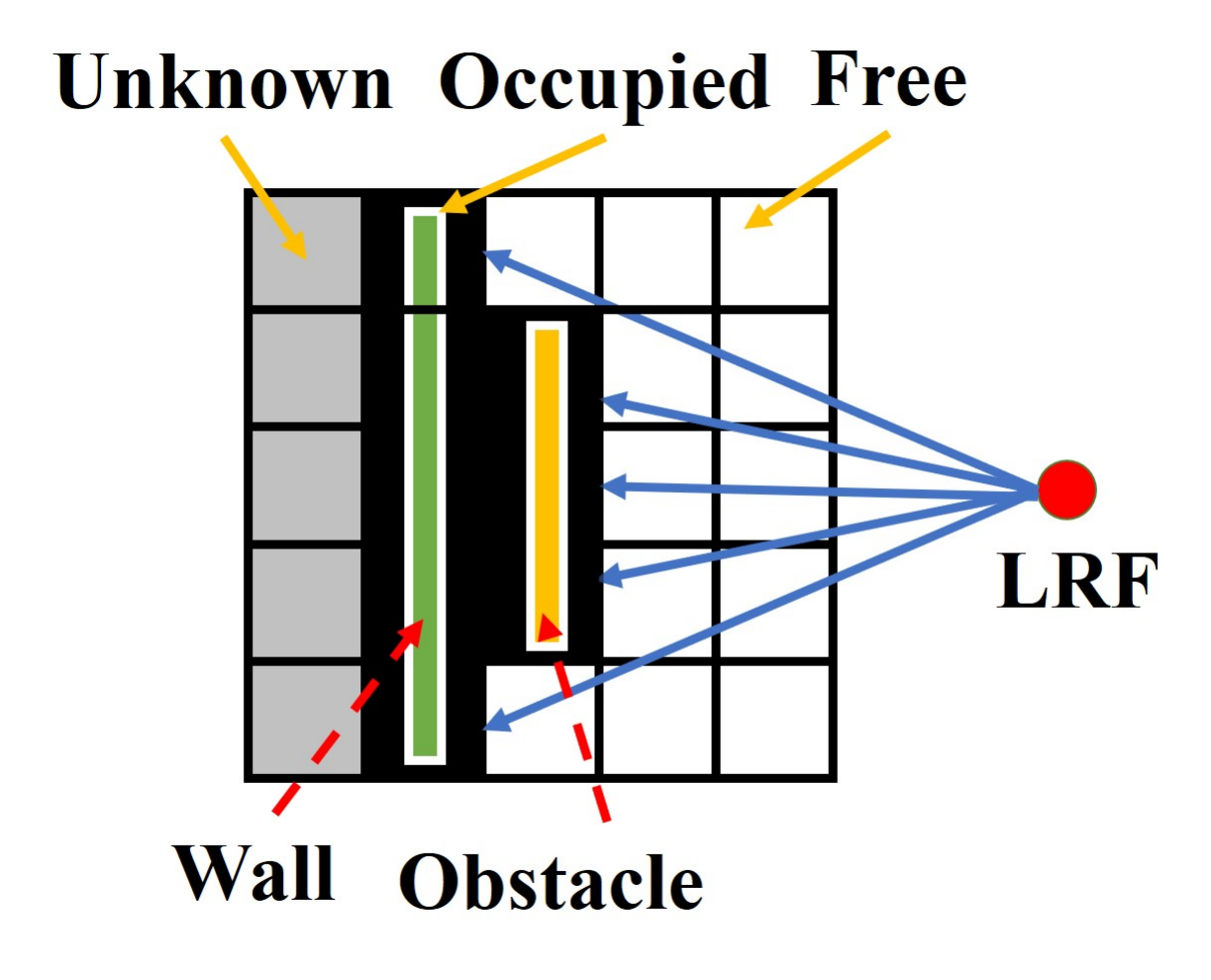}}
      {\begin{center} (c) Obstacle Layer \end{center}}
    \end{center}
  \end{minipage}
  \begin{minipage}{0.49\hsize}
    \begin{center}
      \scalebox{0.16}{
        \includegraphics{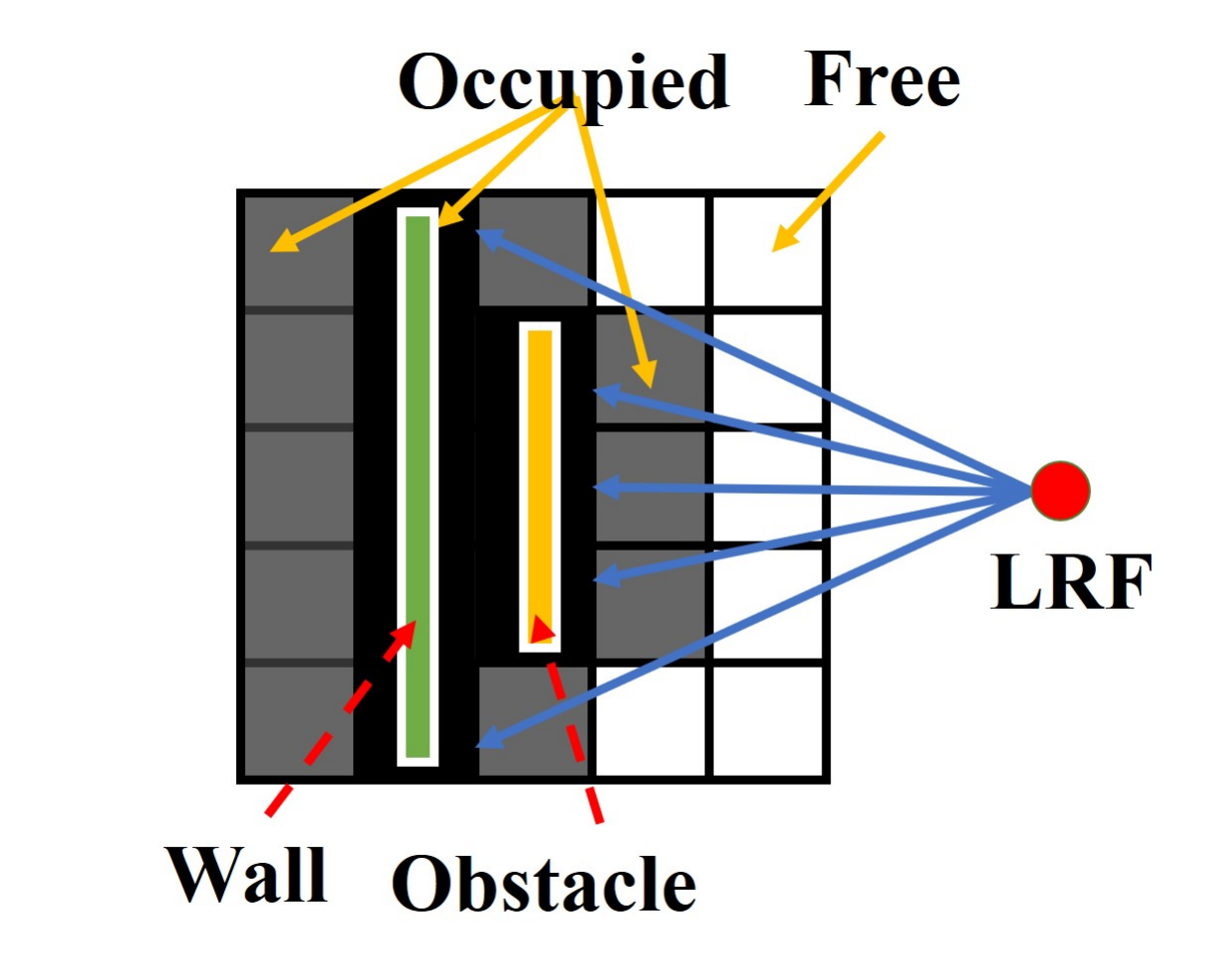}}
      {\begin{center} (d) Inflation Layer \end{center}}
    \end{center}
  \end{minipage}
  \caption{Image Diagram of Layered Cost Map}
  \label{fig:lcm}
\end{figure}

As shown in Fig.~\ref{fig:lcm}, the layered cost map in the ROS navigation stack is applied to the cost function of DWA.
This cost map stores obstacle information obtained from the LRF in three states: ``Free: 0'', ``Occupied: 1-254'' and ``Unknown: 255'' in each divided cell.

In this cost map, three layers are set as the standard in the Layered cost map: ``Static Layer'', ``Obstacle Layer'', and ``Inflation Layer''.
\begin{itemize}
  \item Static Layer: This layer stores the static information of the map generated by the SLAM in advance as shown in Fig.~\ref{fig:lcm}(b).
  \item Obstacle Layer: This layer stores the obstacle data obtained from the distance measurement sensor as shown in Fig.~\ref{fig:lcm}(c).
  \item Inflation Layer: This layer stores the cost of maintaining the safe distance between the robot and the obstacle to prevent the robot from colliding with obstacles as shown in Fig.~\ref{fig:lcm}(d).
\end{itemize}
The path planning is performed in real-time by using (\ref{eq:j_def}) and the cost map as shown in Fig.~\ref{fig:lcm}.

\section{Conventional Method}
This section explains DWA considering blind spots as the conventional method \cite{7}.
By using the cost function with blind spots, the path planning considering the robot's motion performance, collision avoidance, and blind spots were achieved in real-time.
\subsection{Conventional Cost Function}
\begin{figure}[t]
  \begin{minipage}{0.49\hsize}
    \begin{center}
      \scalebox{0.18}{
        \includegraphics{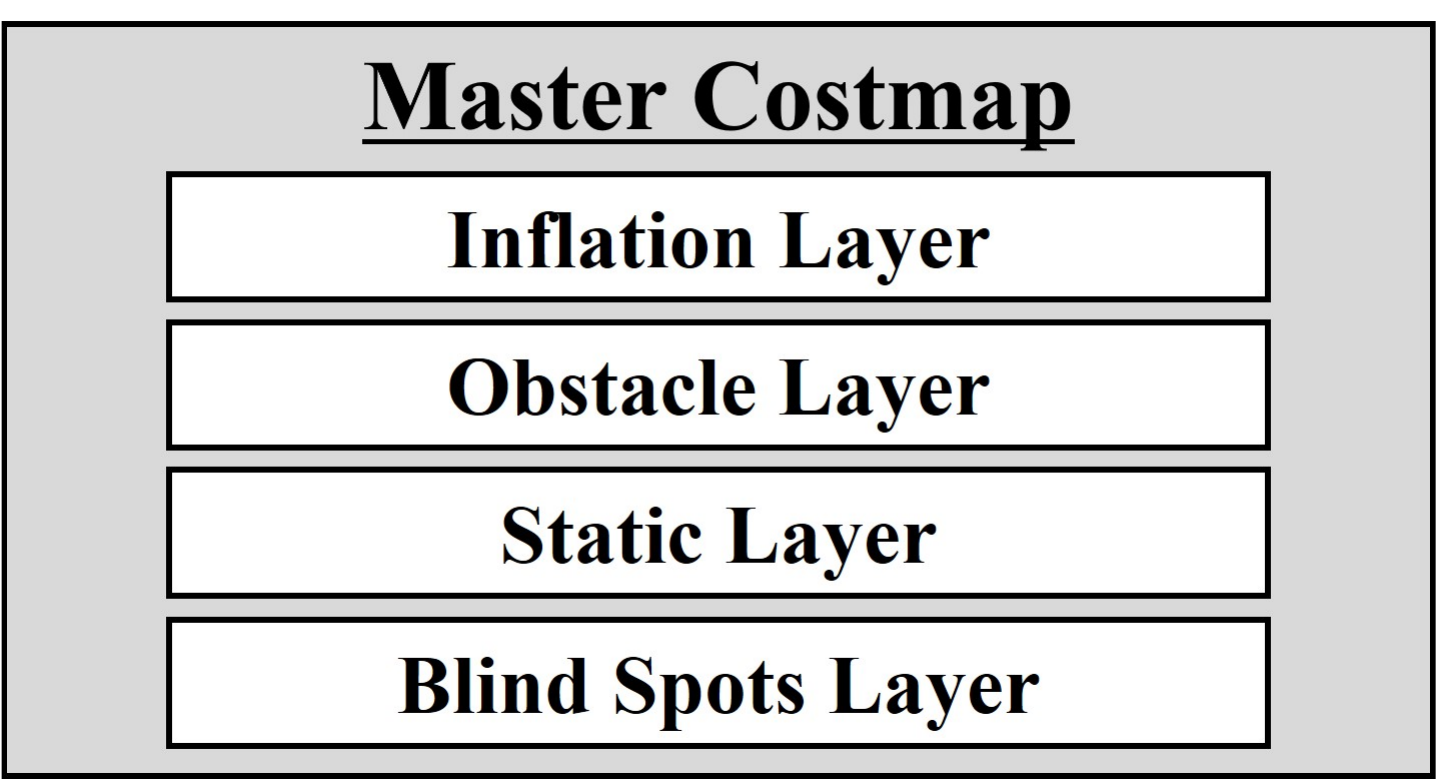}}
      {\begin{center} (a) Master cost map \end{center}}
    \end{center}
  \end{minipage}
  \begin{minipage}{0.49\hsize}
  \vspace{-6mm}
    \begin{center}
      \scalebox{0.22}{
        \includegraphics{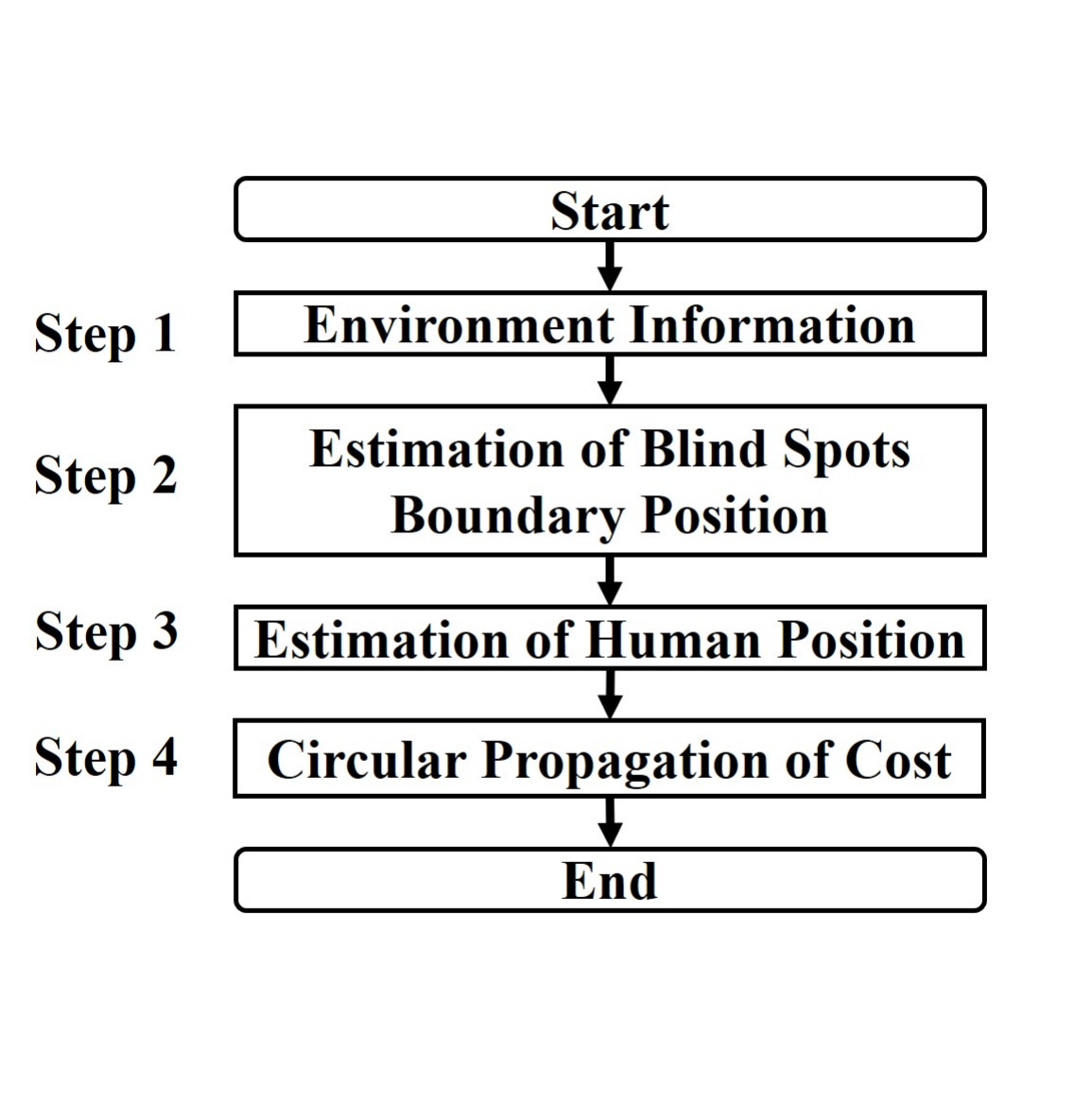}}
      \vspace{-14mm}
      {\begin{center} (b) Flowchart \end{center}}
    \end{center}
  \end{minipage}
  \begin{minipage}{0.49\hsize}
    \begin{center}
     \scalebox{0.31}{
        \includegraphics{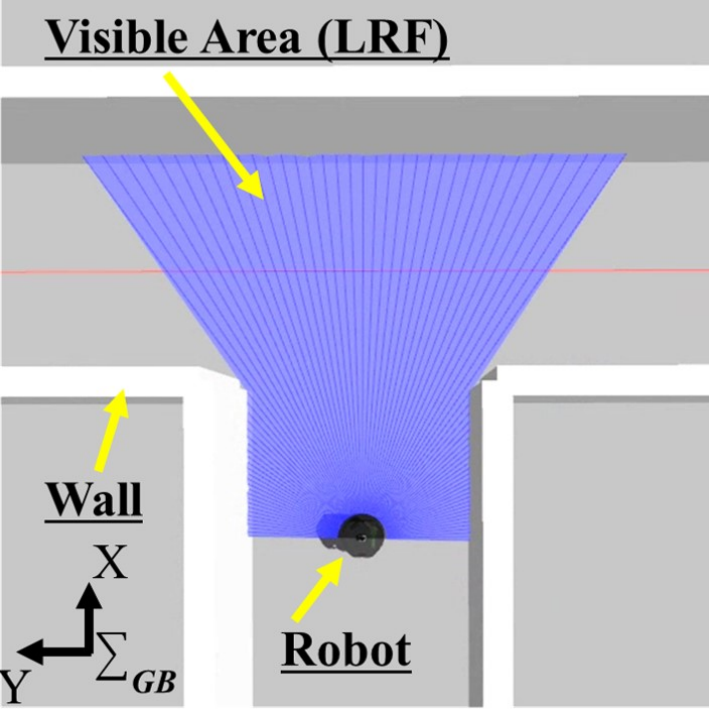}}
      {\begin{center} (c) Step 1 \end{center}}
    \end{center}
  \end{minipage}
  \begin{minipage}{0.49\hsize}
    \begin{center}
     \scalebox{0.31}{
        \includegraphics{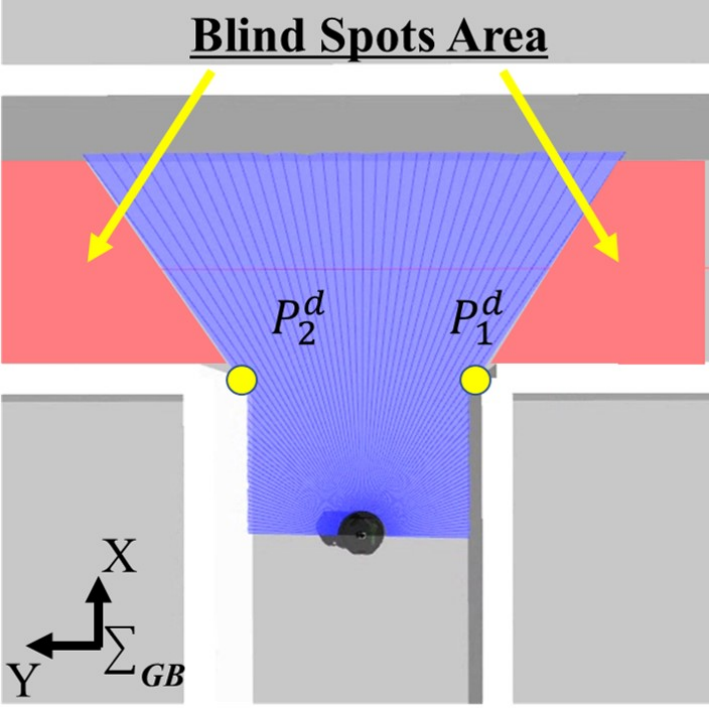}}
      {\begin{center} (d) Step 2 \end{center}}
    \end{center}
  \end{minipage}
  \begin{minipage}{0.49\hsize}
    \begin{center}
     \scalebox{0.31}{
        \includegraphics{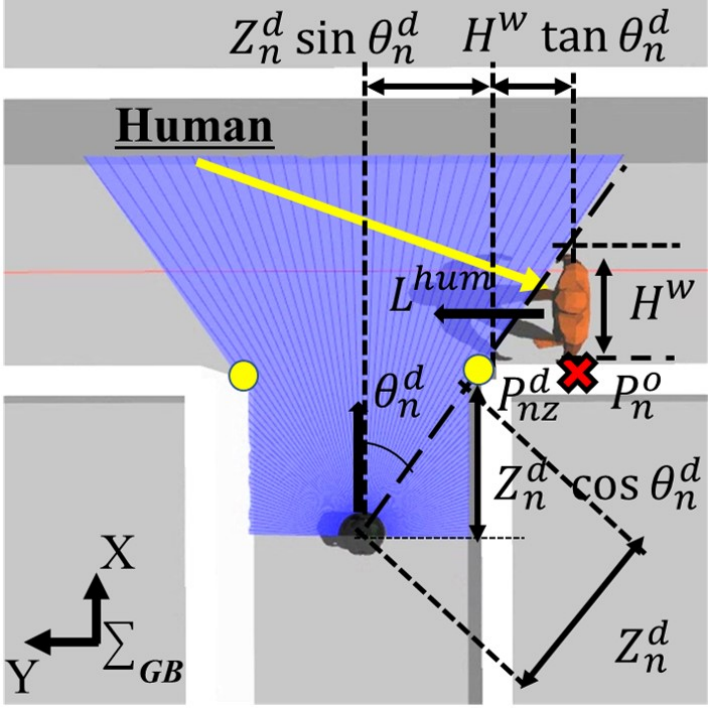}}
      {\begin{center} (e) Step 3 \end{center}}
    \end{center}
  \end{minipage}
  \begin{minipage}{0.49\hsize}
    \begin{center}
     \scalebox{0.31}{
        \includegraphics{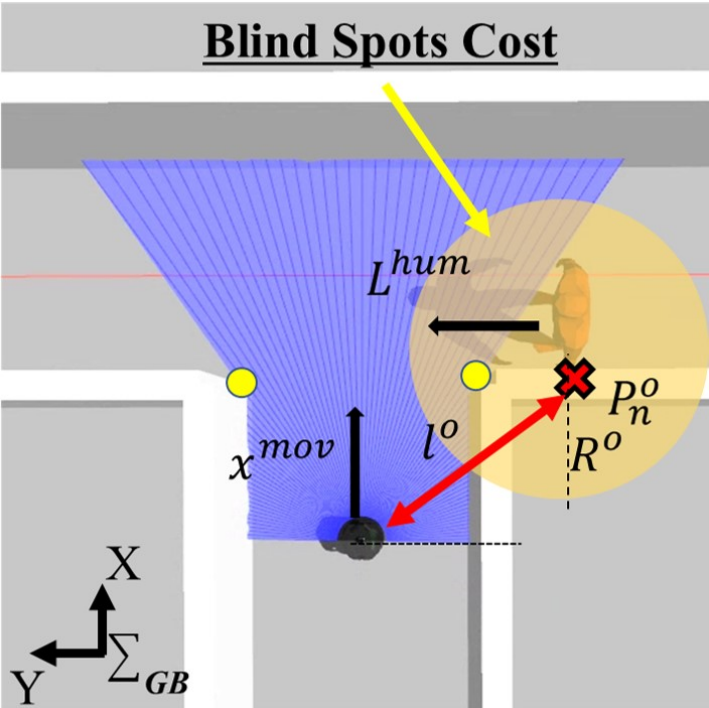}}
      {\begin{center} (f) Step 4 \end{center}}
    \end{center}
  \end{minipage}
  \caption{Coneventional Blind Spots Detection}
  \label{fig:c_bsl}
\end{figure}

The conventional cost function of DWA was defined as (\ref{eq:j_con}).
\begin{equation}
  \label{eq:j_con}
  \begin{split}
    J=W^{pos}\cdot c^{pos}+W^{gol}\cdot c^{gol}+W^{dan}\cdot c^{dan}
  \end{split}
\end{equation}
where $W^{dan}$ represents the weight coefficient considering obstacles and blind spots on the cost map.
$c^{dan}$ represents the maximum map cost considering obstacles and blind spots on the local path.
As shown in Fig.~\ref{fig:c_bsl}(a), the Blind Spots Layer (BSL) is added to the conventional three layers.
By adding the BSL to the cost map system, the path planning takes into account the human and objects coming out of blind spots.
\subsection{Conventional Local Cost Map}
The flowchart shown in Fig.~\ref{fig:c_bsl}(b) is described in detail for each step using Fig.~\ref{fig:c_bsl}(c)-(f).
\subsubsection{Environment Information by LRF}
Fig.~\ref{fig:c_bsl}(c) shows the example of environmental information acquired at the T-intersection.
The sensor measures $i$ ($1 \leq i \leq N$) points as polar coordinates $(Z_i,\theta_i)$.
$N$ means the number of sensor data.
\subsubsection{Estimation of Blind Spots Boundary Position (BSBP)}
Fig.~\ref{fig:c_bsl}(d) shows the conceptual diagram of the blind spots area,
where the red-filled area.
The BSBP ${P^{b}_n}=[Z^{b}_n,\theta^{b}_n]^T$ is defined as the polar coordinate representation in the local coordinate system.
$n$ is the number of BSBP.
The BSBP is calculated from the difference value $(Z^{b}_{i+1}-Z^{b}_i)$ of the neighboring LRF information
which exceeds the threshold value $Z_{th}$.
The BSBP $\bf{P^{b}_n}$ is calculated as follows.
\begin{equation}
  \bf{P^{b}_n}=
  \left(\begin{array}{c}x^{b}_{n}\\ y^{b}_{n}\end{array}\right)=
  \left(\begin{array}{c}Z^{b}_{n}\cos \theta^{b}_{n}\\ Z^{b}_{n} \sin \theta^{b}_{n}\end{array}\right)
  \label{eq:3}
\end{equation}
\begin{figure*}[t]
  \begin{minipage}{0.135\hsize}
    \begin{center}
      \scalebox{0.135}{
        \includegraphics{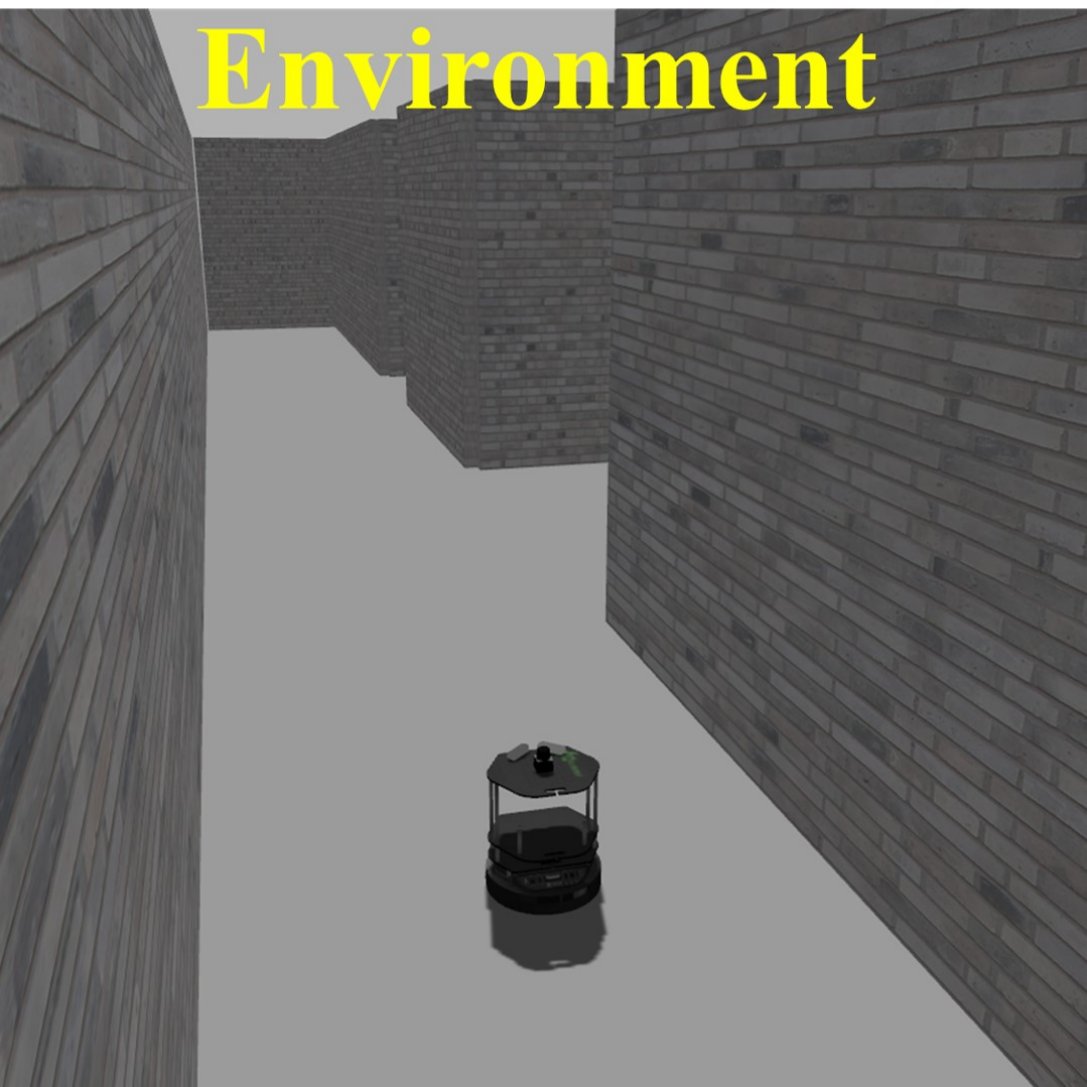}}
      {\begin{center} (a) Environment \end{center}}
    \end{center}
  \end{minipage}
  \begin{minipage}{0.135\hsize}
    \begin{center}
      \scalebox{0.135}{
        \includegraphics{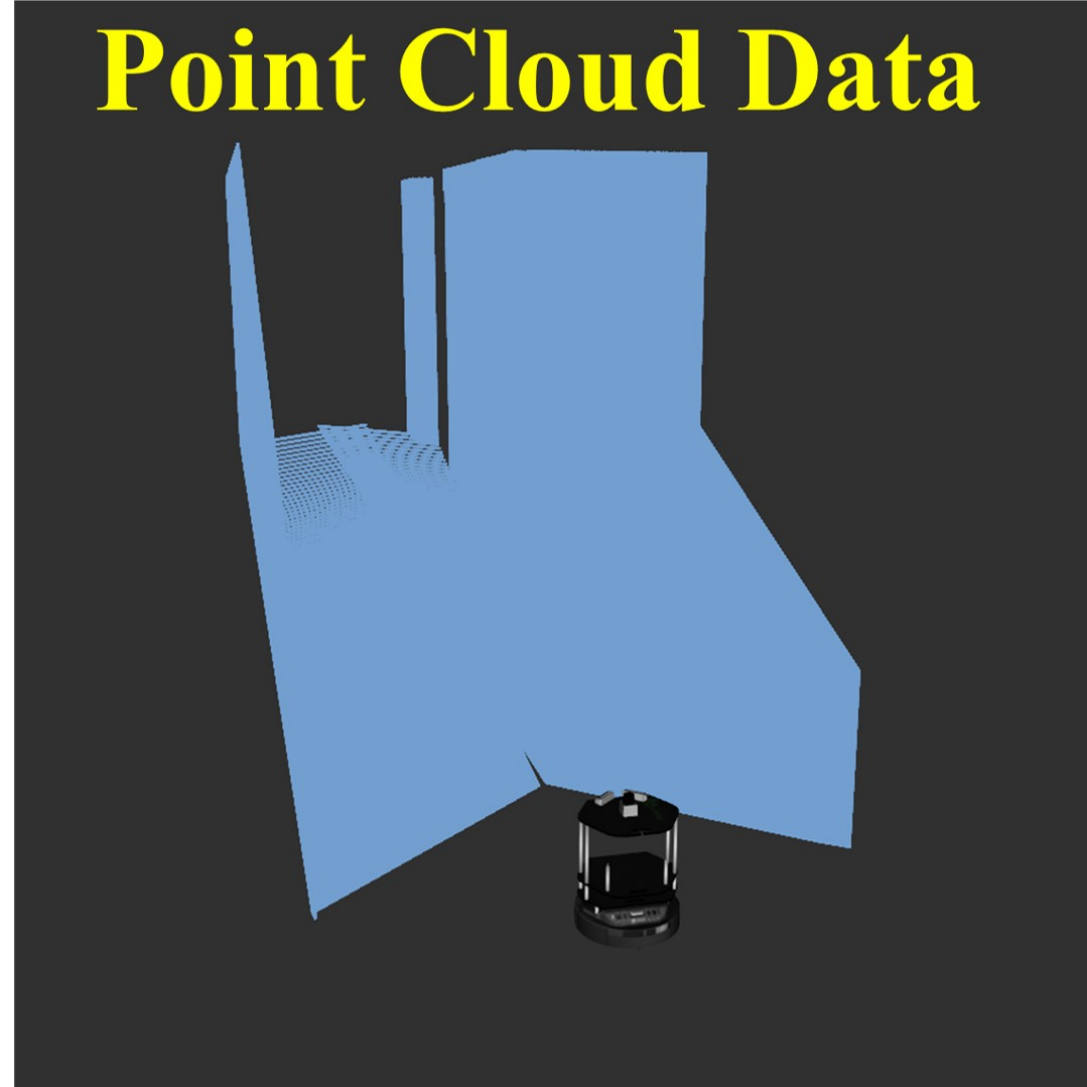}}
      {\begin{center} (b) Step~1 \end{center}}
    \end{center}
  \end{minipage}
  \begin{minipage}{0.135\hsize}
    \begin{center}
      \scalebox{0.135}{
        \includegraphics{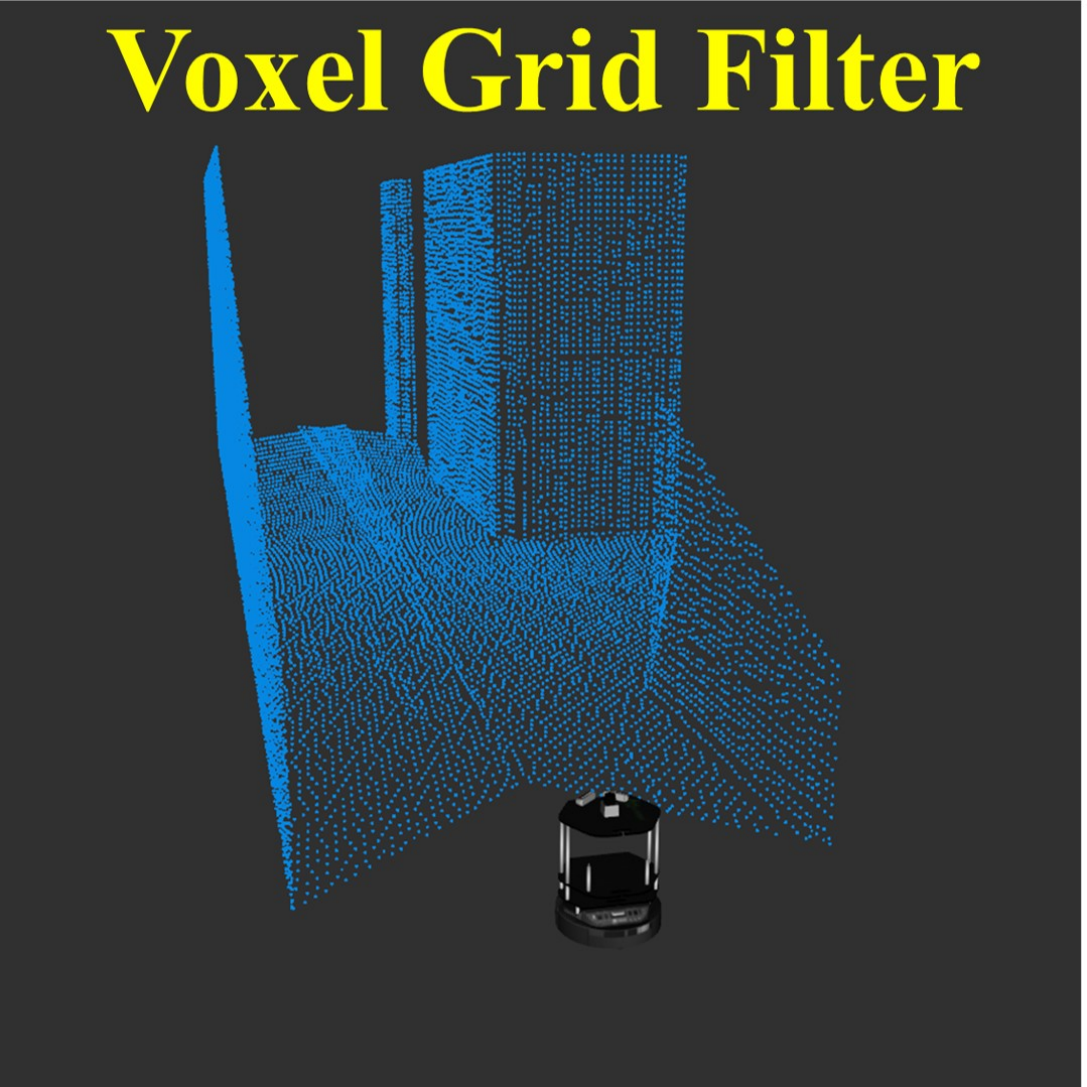}}
      {\begin{center} (c) Step~2a \end{center}}
    \end{center}
  \end{minipage}
  \begin{minipage}{0.135\hsize}
    \begin{center}
      \scalebox{0.135}{
        \includegraphics{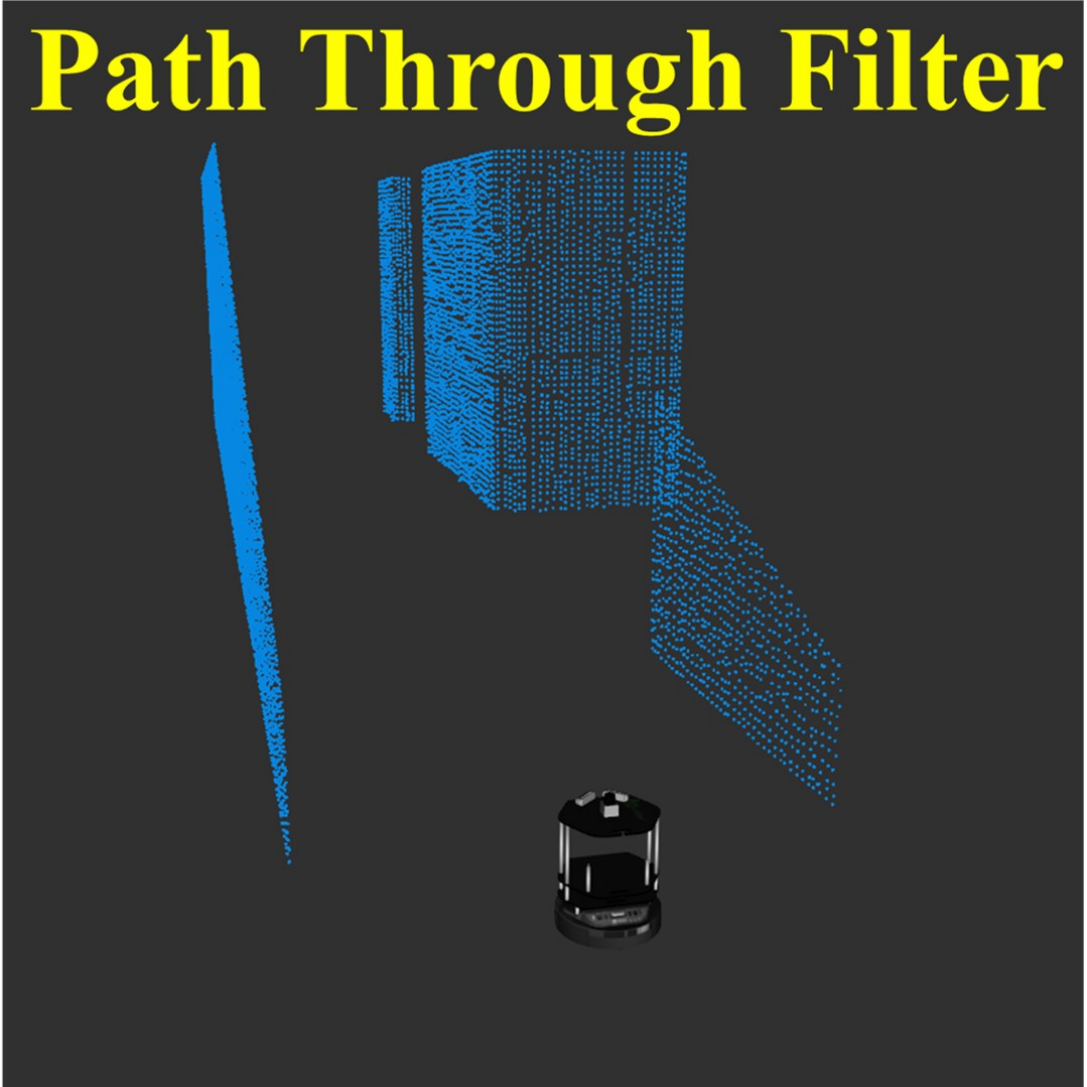}}
      {\begin{center} (d) Step~2b \end{center}}
    \end{center}
  \end{minipage}
  \begin{minipage}{0.135\hsize}
    \begin{center}
      \scalebox{0.135}{
        \includegraphics{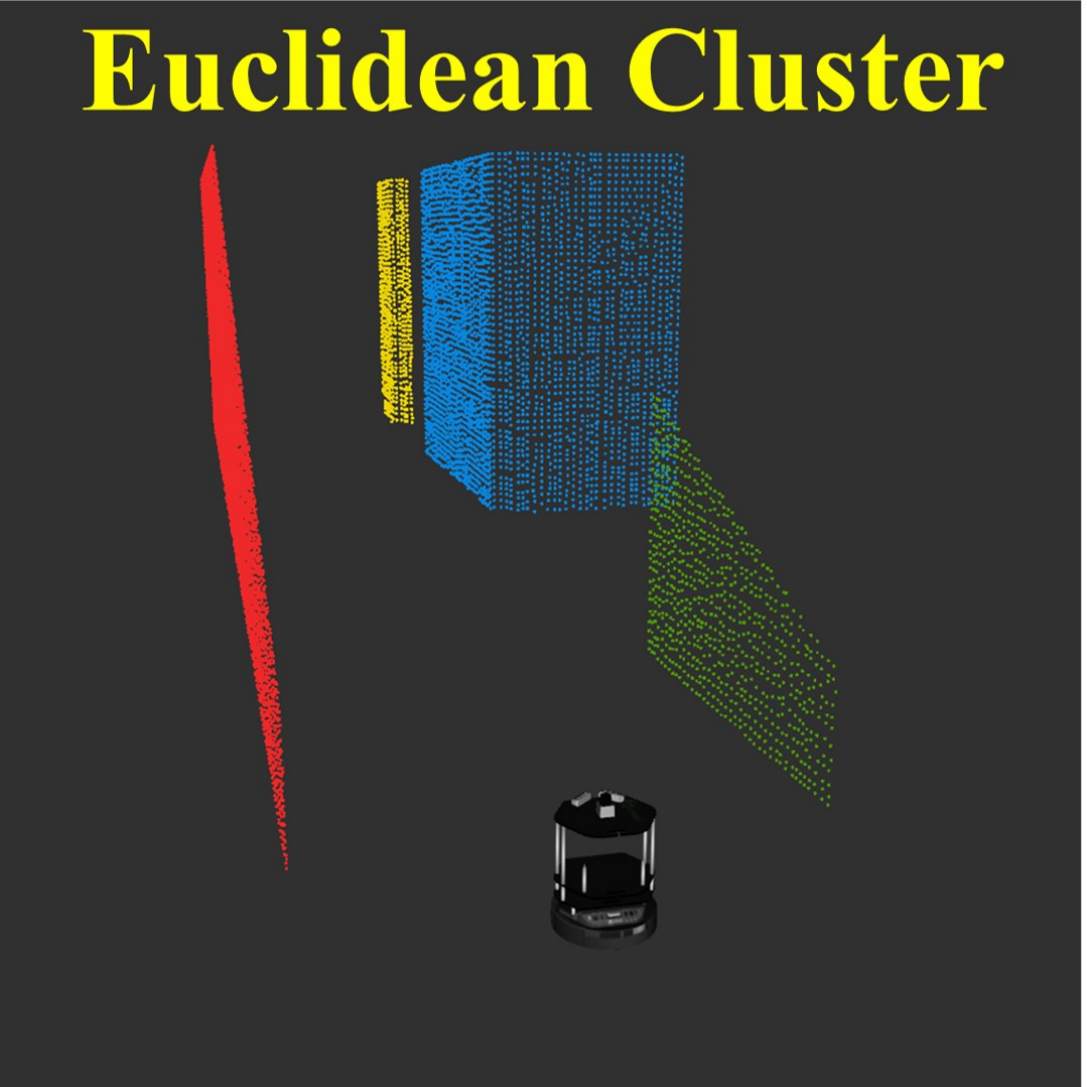}}
      {\begin{center} (e) Step~2c \end{center}}
    \end{center}
  \end{minipage}
  \begin{minipage}{0.135\hsize}
    \begin{center}
      \scalebox{0.135}{
        \includegraphics{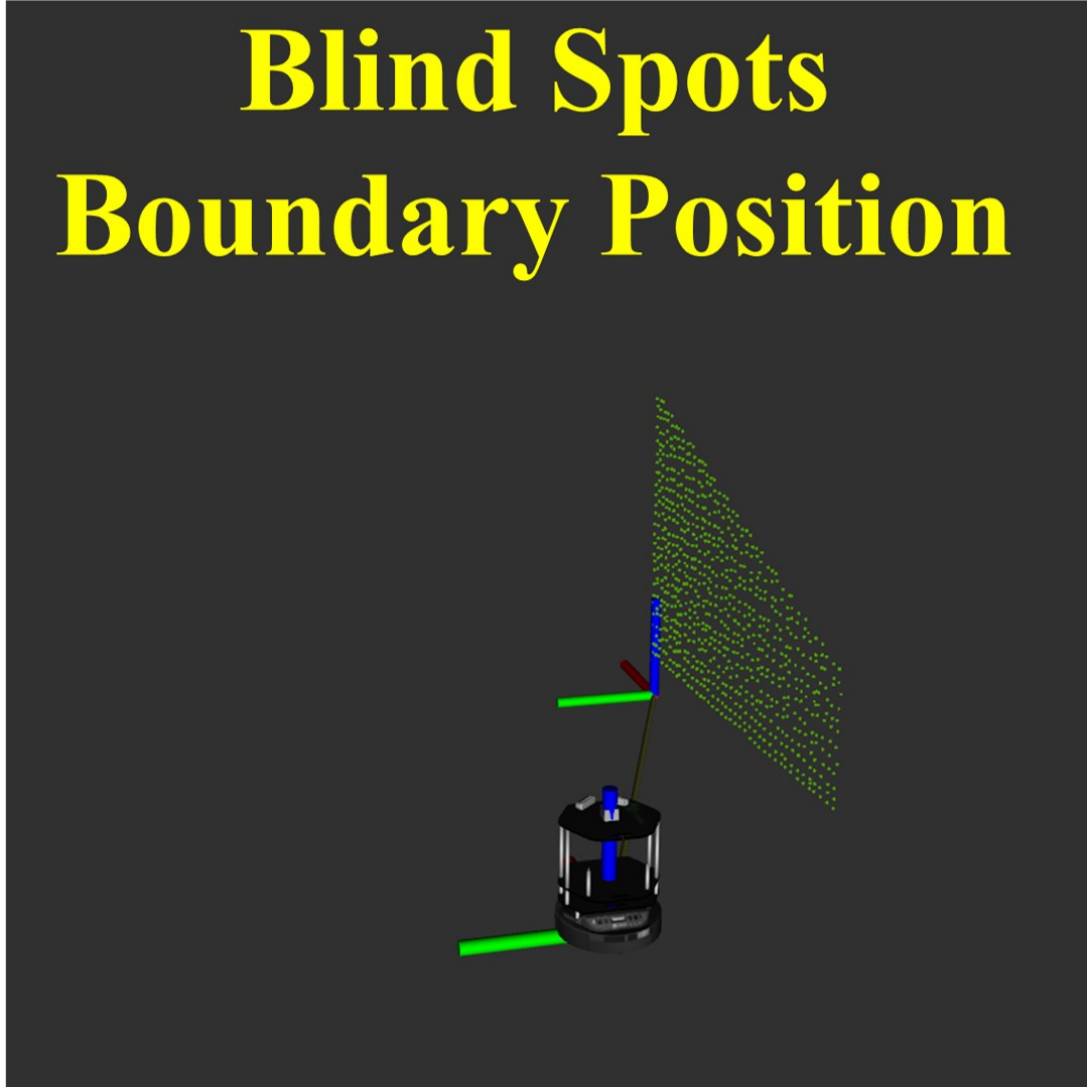}}
      {\begin{center} (f) Step~2d \end{center}}
    \end{center}
  \end{minipage}
  \begin{minipage}{0.135\hsize}
    \begin{center}
      \scalebox{0.135}{
        \includegraphics{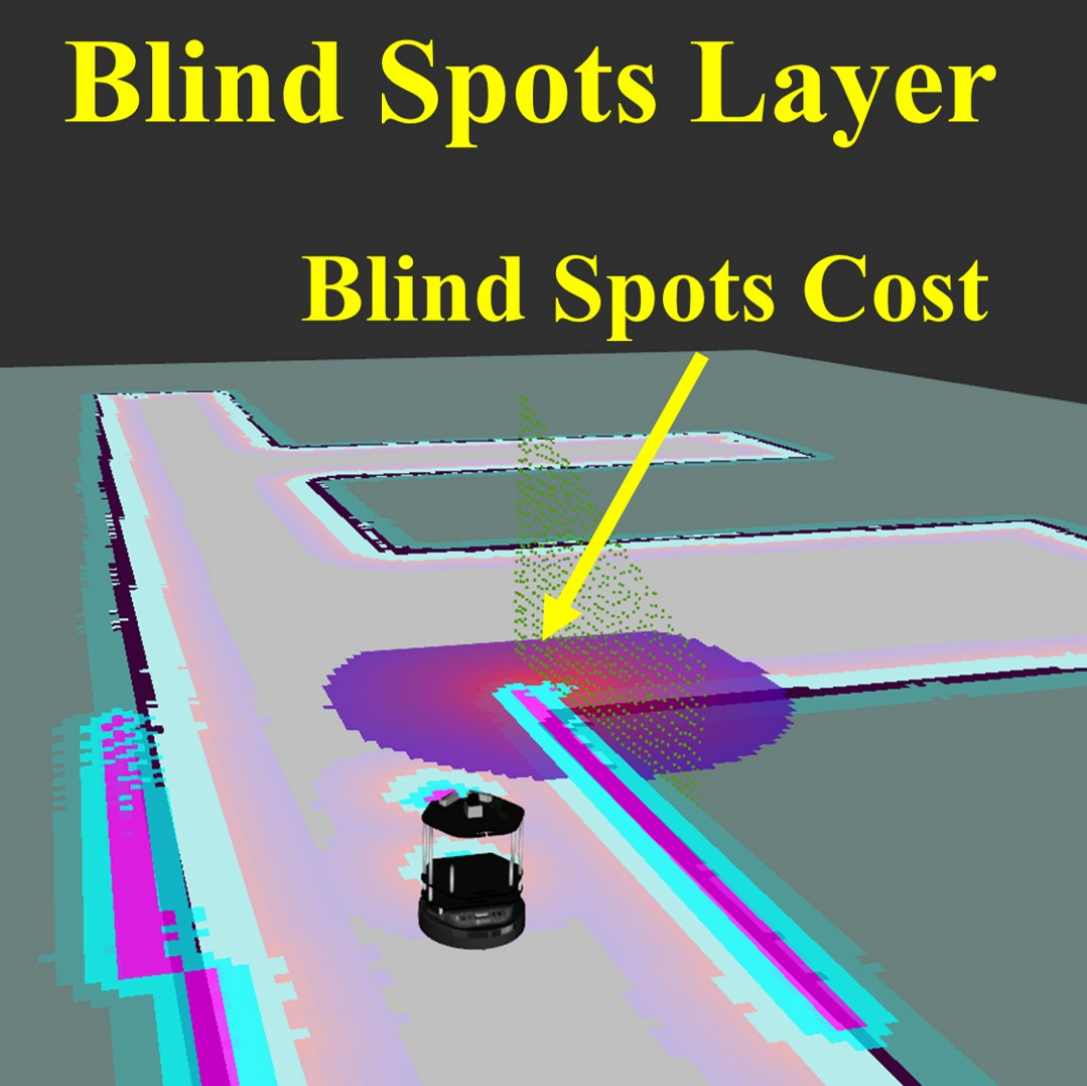}}
      {\begin{center} (g) Step~3-4 \end{center}}
    \end{center}
  \end{minipage}  
  \caption{Propose Blind Spots Detection}
  \label{fig:step_all_bsl}
  \vspace{-4mm}
\end{figure*}
\subsubsection{Estimation of Human Position}
The center of the danger area should be the position closest to the robot in the area where the human may be present.
The center of the dangerous area is calculated from the BSBP ${P^{b}_n}=[x^{b}_n,y^{b}_n]^T$ and the human shoulder-width $H^w$.
Fig.~\ref{fig:c_bsl}(e) shows the center of the danger area.
It is possible to geometrically determine the center of the danger area ${P^o_n}=[x^o_n,y^o_n]^T$ as shown in Fig.~\ref{fig:c_bsl}(e).
The position of the center of the danger area is calculated as follows.
\begin{equation}
  \bf{P^o_n}=
  \left(\begin{array}{c}x^o_{n}\\ y^o_{n}\end{array}\right)=
  \left(\begin{array}{c}x^{b}_{n}\\ y^{b}_{n}+H^w\tan\theta^{b}_{n}\end{array}\right)
  \label{eq:4}
\end{equation}

\subsubsection{Circular Propagation of Cost}
The BSL propagates the cost from the center of the danger area to the cost map in a circular pattern.
It calculates how far to propagate the cost to the cost map for safe path planning based on the stopping distances of the robot and human.
When the robot decelerates with acceleration $a^{mov}$[m/s$^2$] at velocity $v^{mov}$[m/s], the distance for stopping is $x^{mov}$[m], and the time for stopping is $t^{mov}$[sec].
When the robot advances for $t^{mov}$ [sec] until it stops with acceleration $a^{mov}$, the distance $x^{mov}$ is calculated as follows.
\begin{equation}
  x^{mov}=v^{mov}t^{mov}+a^{mov}\frac{(t^{mov})^{2}}{2}
  \label{eq:4.1}
\end{equation}
When the robot decelerates with velocity $v^{mov}$ and acceleration $a^{mov}$, the time for the robot to stop $t^{mov}$ is determined as follows.
\begin{equation}
  t^{mov}=-\frac{v^{mov}}{a^{mov}}
  \label{eq:4.2}
\end{equation}
Substitute equation (\ref{eq:4.2}) into equation (\ref{eq:4.1}) to obtain equation (\ref{eq:4.3}).
\begin{equation}
  x^{mov}=-\frac{(v^{mov})^{2}}{2a^{mov}}
  \label{eq:4.3}
\end{equation}
At the velocity $v^{mov}$, the distance $x^{mov}$ is required for the robot to stop.

The next step is to find the distance until the human stops.
In this paper, it is assumed that human can stop in one step after trying to stop.
Therefore, the stride length of the human is $L^{hum}$[m], which is the distance until the human stop.
As shown in Fig.~\ref{fig:c_bsl}(e)(f), 
$x^{mov}$ means the distance that the robot can stop.
$L^{hum}$ means the distance that the human can stop.
The cost is propagated in the circle from the center of the danger position $P^o_n$ to the distance $R^o$.
\begin{equation}
  R^{o}=x^{mov}+L^{hum}+X^{off}
  \label{eq:4.4}
\end{equation}
where $X^{off}$ is the offset distance, which is set to provide the margin of the distance between the robot and the human.

From the center of the danger area $P^o_n$ to the distance $R^o$, the cost calculated by (\ref{eq:5}) is stored in the cost map.
\begin{equation}
  c^{bsl}=A^{cst}\exp( -S^{cst} l^{dan})
  \label{eq:5}
\end{equation}
where $c^{bsl}$, $S^{cst}$, $l^{dan}$ and $A^{cst}$ represent the cost value determined by the distance to the center of the danger area $P^o_n$, the cost scaling factor, the distance to the center of the dangerous position and the maximum cost value.
\section{Proposed Method}
\subsection{Proposed Cost Function}
In the conventional method\cite{7}, when there are measurement noise of LRF and many small obstacles, the local cost map is filled with blind spot costs.
Therefore, the robot velocity slows down or stops drastically in the situation.
This paper proposes the cost function with the velocity term, so that the robot can achieve the goal without significant deceleration even in the vicinity of blind spot areas.
The cost function of DWA used in the proposed method is as follows.
\begin{equation}
  \begin{split}
    J=W^{pos}\cdot c^{pos}+W^{gol}\cdot c^{gol}+W^{dan}\cdot c^{dan}+W^{vel}\cdot c^{vel}
  \end{split}
  \label{eq:j_pro}
\end{equation}
where $W^{vel}$ and $c^{vel}$ represent the weight coefficient considering translational velocity and the reciprocal of the current translational velocity.

\subsection{Proposed Local Cost Map}
The LRF is used for blind spots detection in the conventional method.
The blind spots detection range is limited to the horizontal plane of the LRF,  which is not flexible enough for various environments.
In the proposed method, RGB-D cameras are used to calculate the blind spots detection.
As shown in Fig.~\ref{fig:c_bsl}(b) the proposed method is similar to the conventional method except for the {\it Step 1} and {\it Step 2}.
The point cloud information acquired from RGB-D cameras in {\it Step 1} is used to calculate the BSBP.
This section describes the difference {\it Step 2} between the proposed and conventional methods.
\subsubsection{Voxel Grid Filter (Step 2a)}

As shown in Fig.~\ref{fig:step_all_bsl}(a)-(c), the robot accrued point cloud data from RGB-D cameras.
The space of the point cloud is delimited by voxels, and points are approximated by the point cloud center of gravity in each voxel.
The number of points is reduced, and the computational cost is reduced.
\subsubsection{Path Through Filter (Step 2b)}
As shown in Fig.~\ref{fig:step_all_bsl}(d), the path through the filter removes the point cloud of the ground.
\subsubsection{Euclidean Cluster Extraction (Step 2c)}
As shown in Fig.~\ref{fig:step_all_bsl}(e), the clusters of point clouds where the distance between points is less than or equal to threshold values are considered to be the same cluster.
\subsubsection{Blind Spots Boundary Position (Step 2d)}
The robot extracts the nearest left and right point cluster as shown in Fig.~\ref{fig:step_all_bsl}(f).
BSBP $\bf{P^{b}_n}$ is calculated from the maximum value of the X-axis and the maximum and minimum values of the Y-axis of the point cluster in the local coordinate system.
The proposed method defines BSBP as the boundary of the observable point cloud.

\begin{equation}
  \bf{P^{b}_n}=
  \left(\begin{array}{c}
      x^{b}_{n} \\
      y^{b}_{n}\end{array}
  \right)=\left(\begin{array}{c}\arg\max(\bf{\Gamma}^x_n) \\
      \frac{\arg\max(\bf{\Gamma}^y_n)+\arg\min(\bf{\Gamma}^y_n)}{2}\end{array}\right)
  \label{eq:7}
\end{equation}
where $\bf{\Gamma}^x_n$ is the $X$-coordinate value of point cloud in the $n$-th cluster and $\bf{\Gamma}^y_n$ is the $Y$-coordinate value of point cloud in the $n$-th cluster.

In the proposed method, the {\it Step 3} and {\it 4} are performed using (\ref{eq:7}), and the cost is generated as shown in Fig.~\ref{fig:step_all_bsl}(g).
\subsection{Example of Proposed Method}
\begin{figure*}[t]
  \begin{minipage}{0.245\hsize}
    \begin{center}
      \scalebox{0.42}{
        \includegraphics{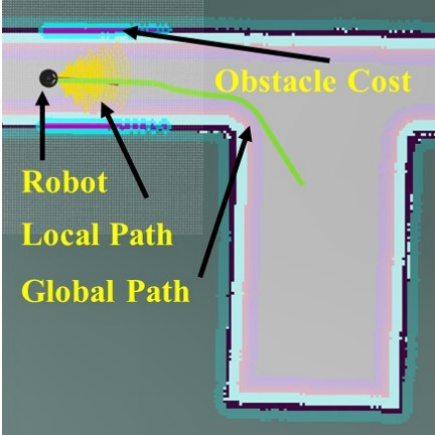}}
      \vspace{-2mm}
      {\begin{center} (a) Scene 1\end{center}}
    \end{center}
  \end{minipage}
  \begin{minipage}{0.245\hsize}
    \begin{center}
      \scalebox{0.42}{
        \includegraphics{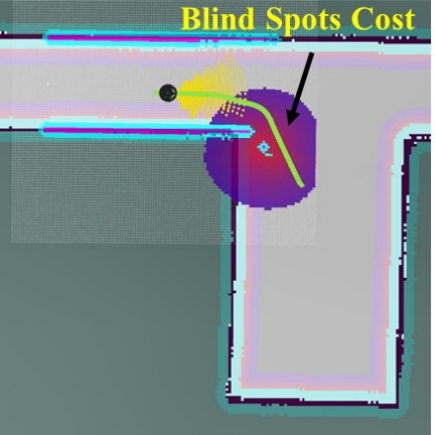}}
      \vspace{-2mm}
      {\begin{center} (b) Scene 2\end{center}}
    \end{center}
  \end{minipage}
  \begin{minipage}{0.245\hsize}
    \begin{center}
      \scalebox{0.42}{
        \includegraphics{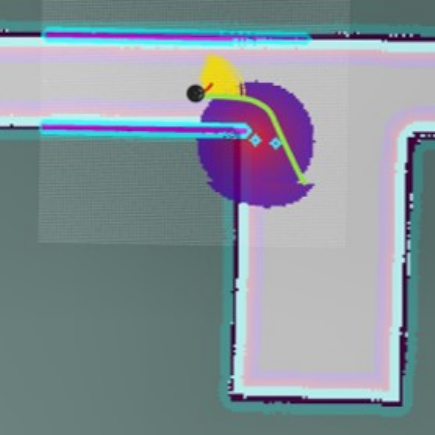}}
      \vspace{-2mm}
      {\begin{center} (c) Scene 3\end{center}}
    \end{center}
  \end{minipage}
  \begin{minipage}{0.245\hsize}
    \begin{center}
      \scalebox{0.42}{
        \includegraphics{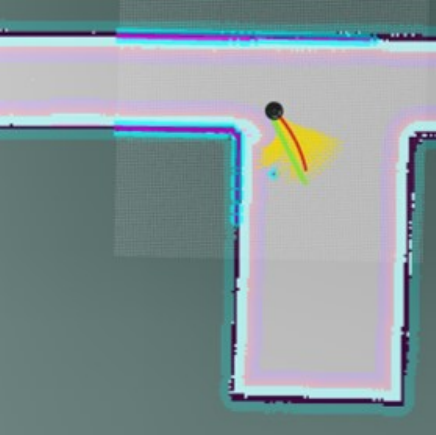}}
        \vspace{-2mm}
        {\begin{center} (d) Scene 4\end{center}}
    \end{center}
  \end{minipage}
  \caption{Example of Proposed Method}
  \label{fig:cost}
  \vspace{-4mm}
\end{figure*}

Fig.~\ref{fig:cost} shows an example of the proposed method.
The green line is the path calculated by global path planning.
The yellow fan-shaped lines are the path candidates of DWA.
Furthermore, the red bold line is the optimal path determined from DWA.
The robot uses the red bold line as the command value of velocities.
There are no blind spots in the local cost map, so DWA does not take blind spots into account (Fig.~\ref{fig:cost}(a)).
The blind spot area is detected by RGB-D cameras. The cost is propagated in a circle (Fig.~\ref{fig:cost}(b)).
The red line of DWA is selected to avoid the blind spot area (Fig.~\ref{fig:cost}(c)).
The blind spot area is eliminated and the local path is selected to follow the global path plan(Fig.~\ref{fig:cost}(d)).

\section{Simulation}
\subsection{Simulation Setup}
\subsubsection{Simulation Environment}

\begin{figure*}[t]
  \begin{minipage}{0.33\hsize}
    \begin{center}
      \scalebox{0.23}{
        \includegraphics{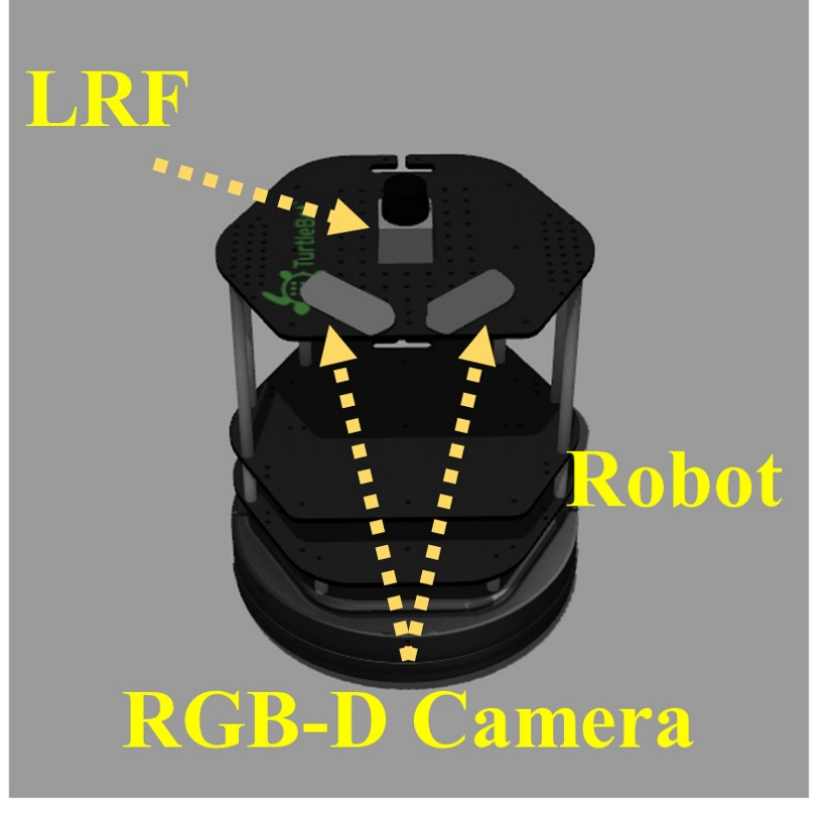}}
      \vspace{-2mm}
      {\begin{center} (a) Robot \end{center}}
    \end{center}
  \end{minipage}
  \begin{minipage}{0.33\hsize}
    \begin{center}
      \scalebox{0.35}{
        \includegraphics{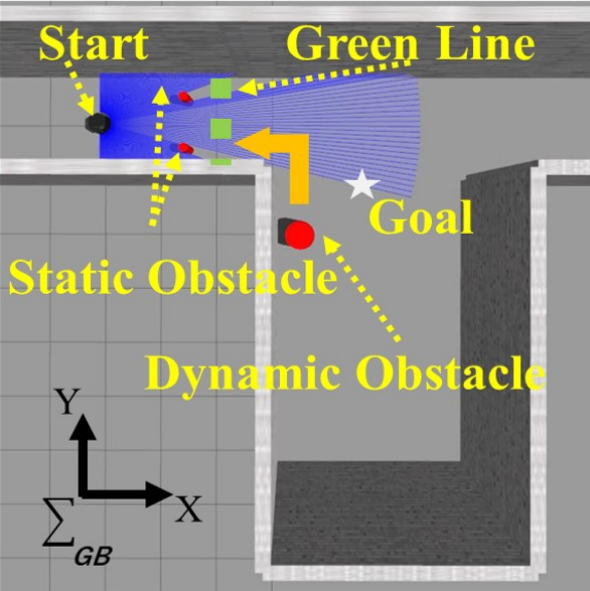}}
      \vspace{-2mm}
      {\begin{center} (b) Environment (Case S1)\end{center}}
    \end{center}
  \end{minipage}
  \begin{minipage}{0.33\hsize}
    \begin{center}
      \scalebox{0.35}{
        \includegraphics{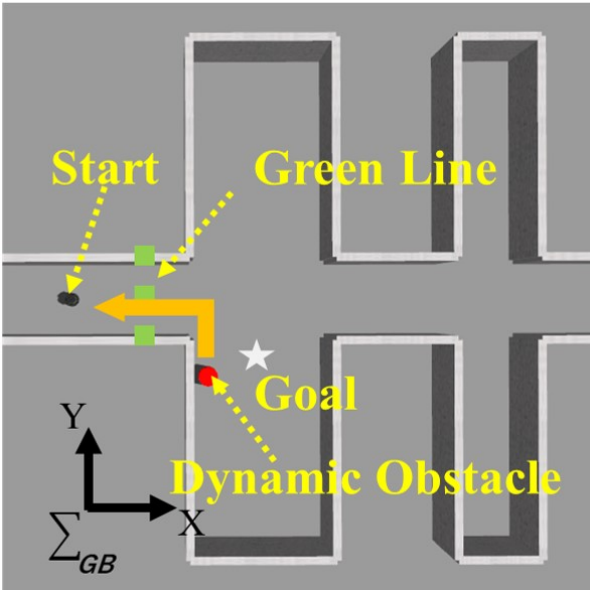}}
      \vspace{-2mm}
      {\begin{center} (c) Environment (Case S2)\end{center}}
    \end{center}
  \end{minipage}  
  \caption{Simulation Environment}
  \label{fig:se}
  \vspace{-4mm}
\end{figure*}

\begin{table}[t]
  \begin{center}
    \caption{Experimental Parameters}
    \scalebox{0.8}{
      \begin{tabular}{|c|c|c|} \hline
        Character  & Value     & Description                                             \\ \hline\hline
        $L$        & 0.8[m]    & Human Stride                                            \\\hline
        $X^{off}$   & 0.2[m]    & Offset Distance                                         \\\hline
        $H^w$      & 0.5[m]    & Human Shoulders                                         \\\hline
        $S^{cst}$ & 1         & Cost Scaling Factor                                     \\\hline
        $A^{cst}$ & 253       & Maximum Cost                                            \\\hline
        $W^pos$      & 2         & Weight Coefficient for Global Path                      \\\hline
        $W^{gol}$      & 1         & Weight Coefficient for Goal Position                    \\\hline
        $W^{obs}$      & 10        & Weight Coefficient for Obstacles                        \\\hline
        $W^{ban}$      & 10        & Weight Coefficient for Obstacles and Blind Spots Region \\\hline
        $W^{vel}$      & 0.5        & Weight Coefficient for Velocity \\\hline
        $T^{pre}$  & 4.0[sec]    & Predicted Time                                          \\\hline
        $Z^{thr}$   & 1.0       & Threshold of BSBP              \\\hline
      \end{tabular}
    }
    \label{1}
  \end{center}
\end{table}

\begin{table}[t]
  \begin{center}
    \caption{Simulation Setup}
    \scalebox{0.8}{
      \begin{tabular}{|l|l|l|l|} \hline
        Method   & Cost Map                          & Global / Local Planner & Cost Function \\\hline
        Method~1 & ROS Default (Fig.~\ref{fig:lcm}) & A* / DWA               & eq.~(\ref{eq:j_def})     \\\hline
        Method~2 & ROS Default + BSL (LRF)           & A* / DWA               & eq.~(\ref{eq:j_con})     \\\hline
        Method~3 & ROS Default + BSL (RGB-D)         & A* / DWA               & eq.~(\ref{eq:j_con})     \\\hline
        Method~4 & ROS Default + BSL (RGB-D)         & A* / DWA               & eq.~(\ref{eq:j_pro})     \\\hline
      \end{tabular}
    }
    \label{tab_method}
  \end{center}
\end{table}
\begin{table}[t]
  \begin{center}
    \caption{Simulation Results Case S1}
    \scalebox{0.85}{
      \begin{tabular}{|l|l|l|l|} \hline
        Navigation Method                    &Cost Function& Goal       & Time [sec] \\\hline
        Method~1 : ROS Default               &eq.~(\ref{eq:j_def})& $\times $  & -          \\\hline
        Method~2 : ROS Default + BSL (LRF)   &eq.~(\ref{eq:j_con})& $\bigcirc$ & 25.3       \\\hline
        Method~3 : ROS Default + BSL (RGB-D) &eq.~(\ref{eq:j_con})& $\bigcirc$ & 20.2       \\\hline
        Method~4 : ROS Default + BSL (RGB-D) &eq.~(\ref{eq:j_pro})& $\bigcirc$ & 18.3       \\\hline
      \end{tabular}
    }
    \label{tab2}
  \end{center}
\end{table}
\begin{table}[t]
  \begin{center}
    \caption{Simulation Results Case S2}
    \scalebox{0.85}{
      \begin{tabular}{|l|l|l|l|} \hline
        Navigation Method                    &Cost Function& Goal       & Time [sec] \\\hline
        Method~1 : ROS Default               &eq.~(\ref{eq:j_def})& $\times $  & -          \\\hline
        Method~2 : ROS Default + BSL (LRF)   &eq. (\ref{eq:j_con})& $\bigcirc$ & 23.8       \\\hline
        Method~3 : ROS Default + BSL (RGB-D) &eq. (\ref{eq:j_con})& $\bigcirc$ & 21.2       \\\hline
        Method~4 : ROS Default + BSL (RGB-D) &eq. (\ref{eq:j_pro})& $\bigcirc$ & 20.5       \\\hline
      \end{tabular}
    }
    \label{sim-res-rgbd2}
  \end{center}
\end{table}

\begin{figure*}[t]
  \begin{minipage}{0.245\hsize}
    \begin{center}
      \scalebox{0.18}{
      \includegraphics{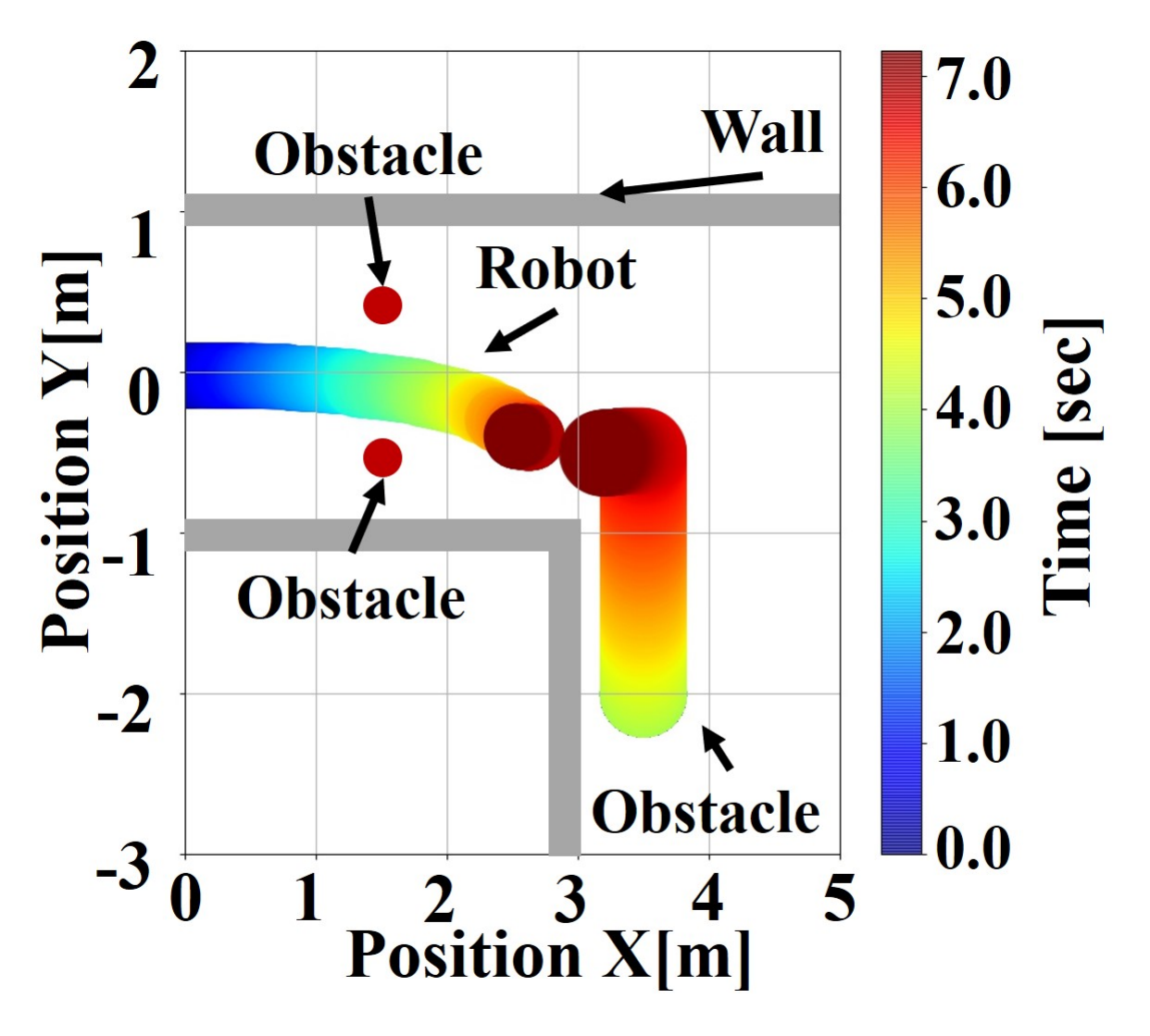}}
   \vspace{-5mm}
      {\begin{center} (a) {\it Method 1} \end{center}}
    \end{center}
  \end{minipage}
  \begin{minipage}{0.245\hsize}
    \begin{center}
      \scalebox{0.18}{
      \includegraphics{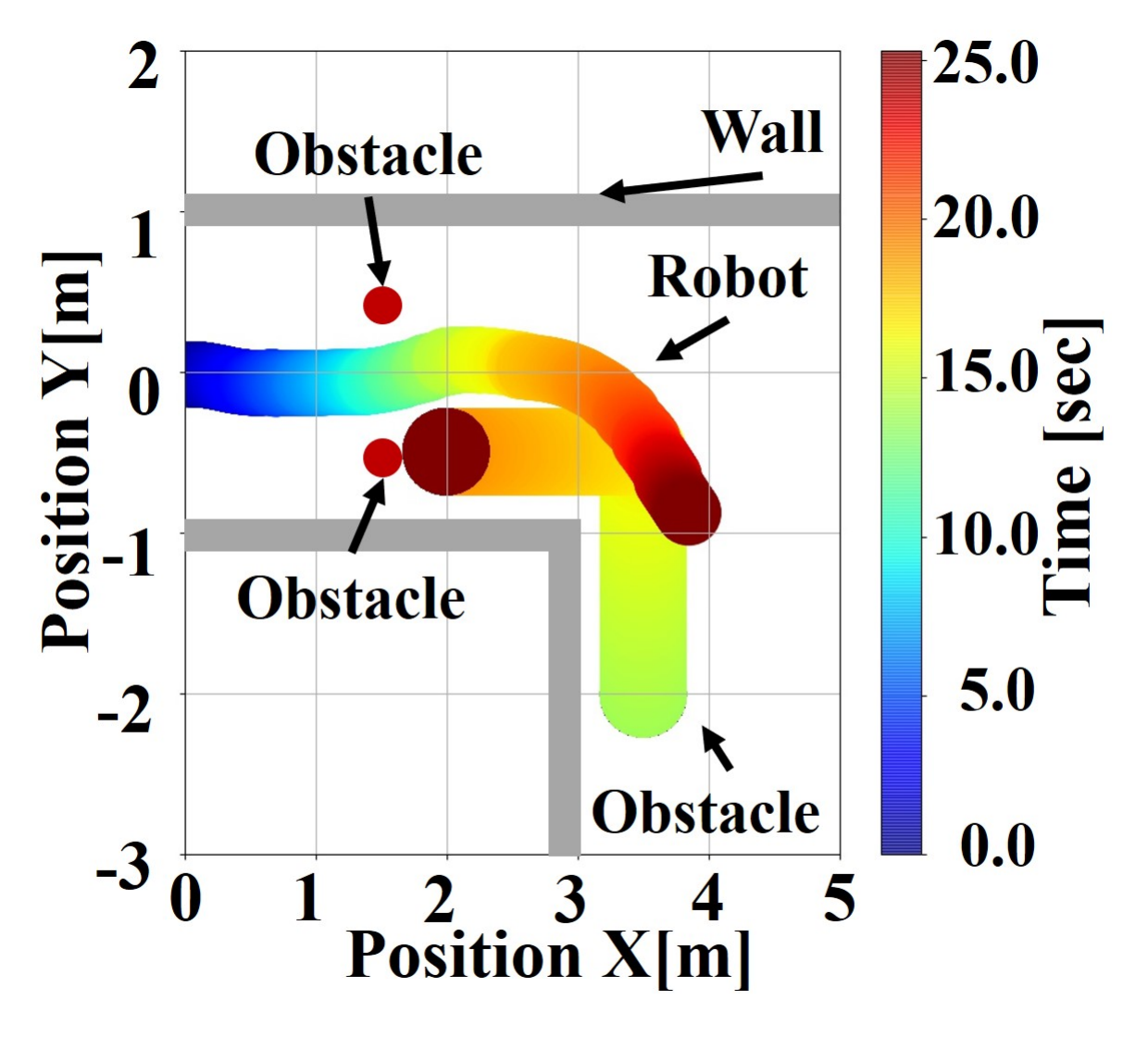}}
   \vspace{-5mm}
      {\begin{center} (b) {\it Method 2} \end{center}}
    \end{center}
  \end{minipage}
  \begin{minipage}{0.245\hsize}
    \begin{center}
      \scalebox{0.25}{
      \includegraphics{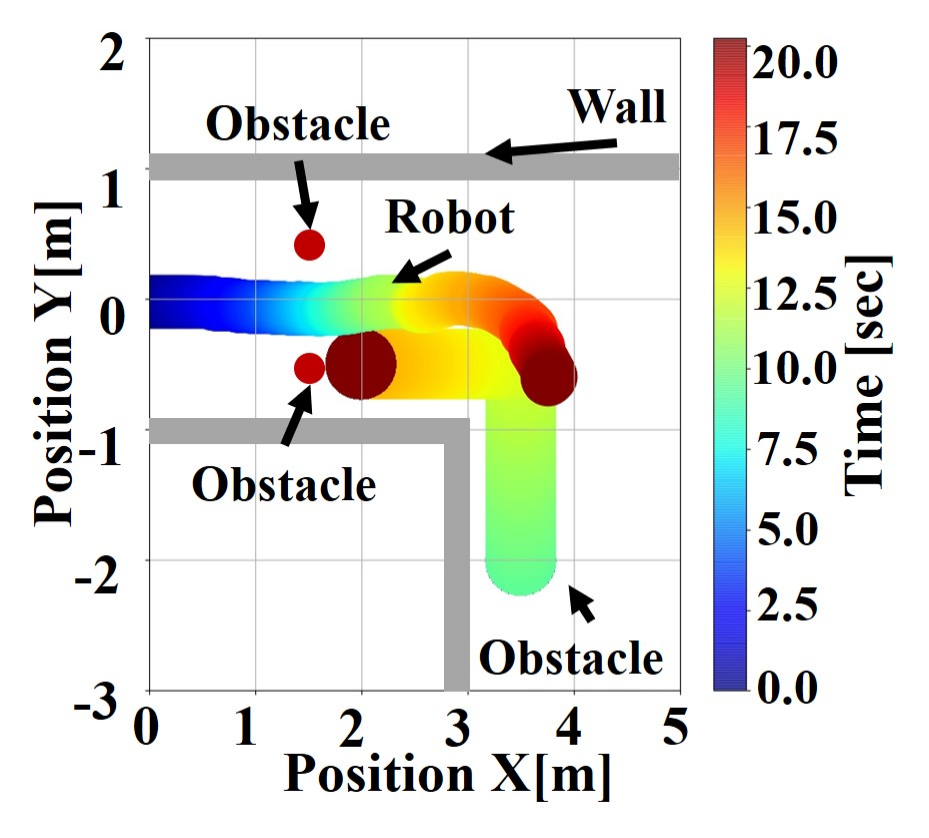}}
   \vspace{-5mm}
      {\begin{center} (c) {\it Method 3} \end{center}}
    \end{center}
  \end{minipage}
  \begin{minipage}{0.245\hsize}
    \begin{center}
      \scalebox{0.18}{
      \includegraphics{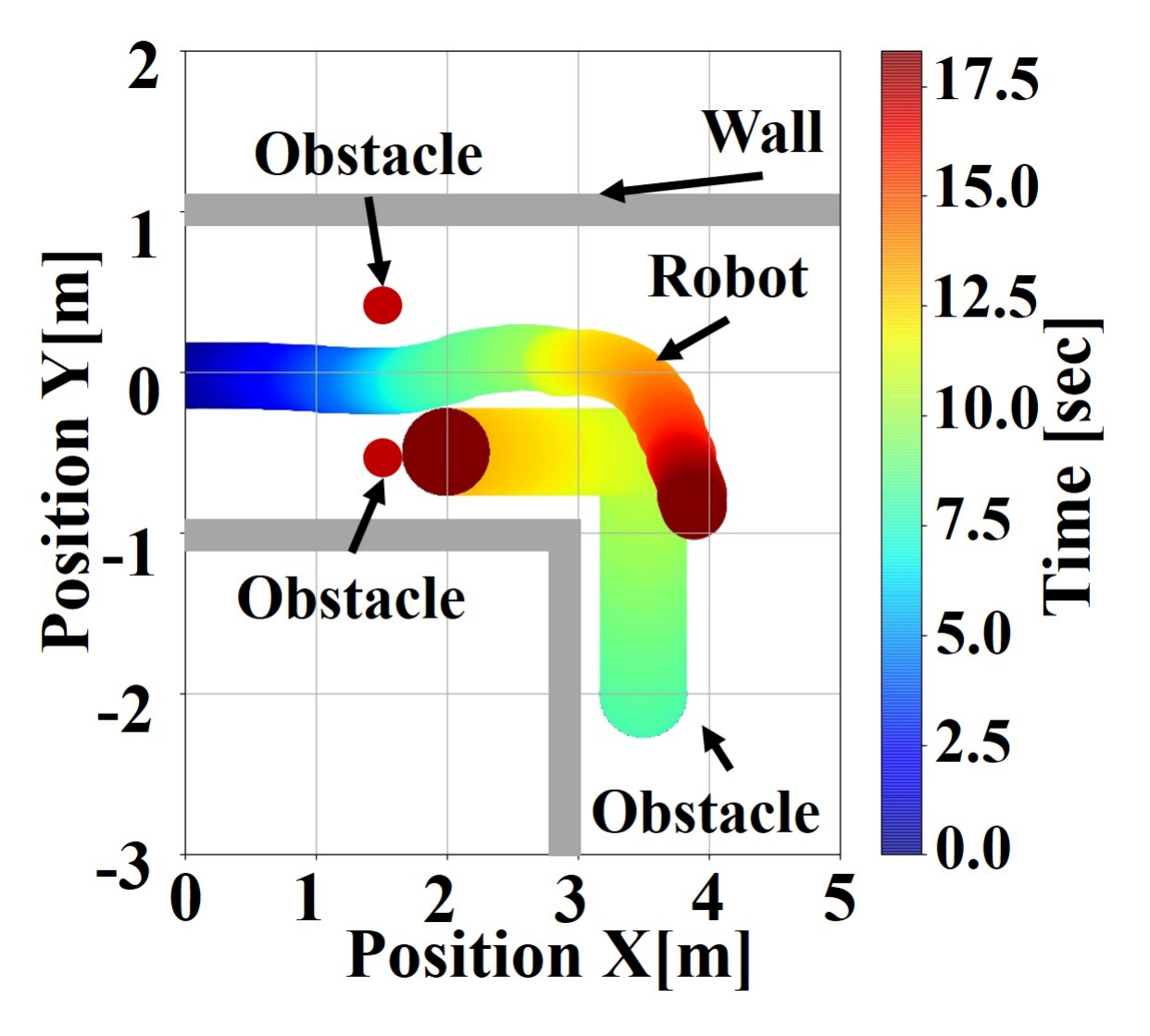}}
   \vspace{-5mm}
      {\begin{center} (d) {\it Method 4} \end{center}}
    \end{center}
  \end{minipage}
  \caption{Simulation Results (Case S1)}
  \label{fig:sim-res-rgbd1}
\end{figure*}

\begin{figure*}[t]
  \begin{minipage}{0.245\hsize}
    \begin{center}
      \scalebox{0.25}{
      \includegraphics{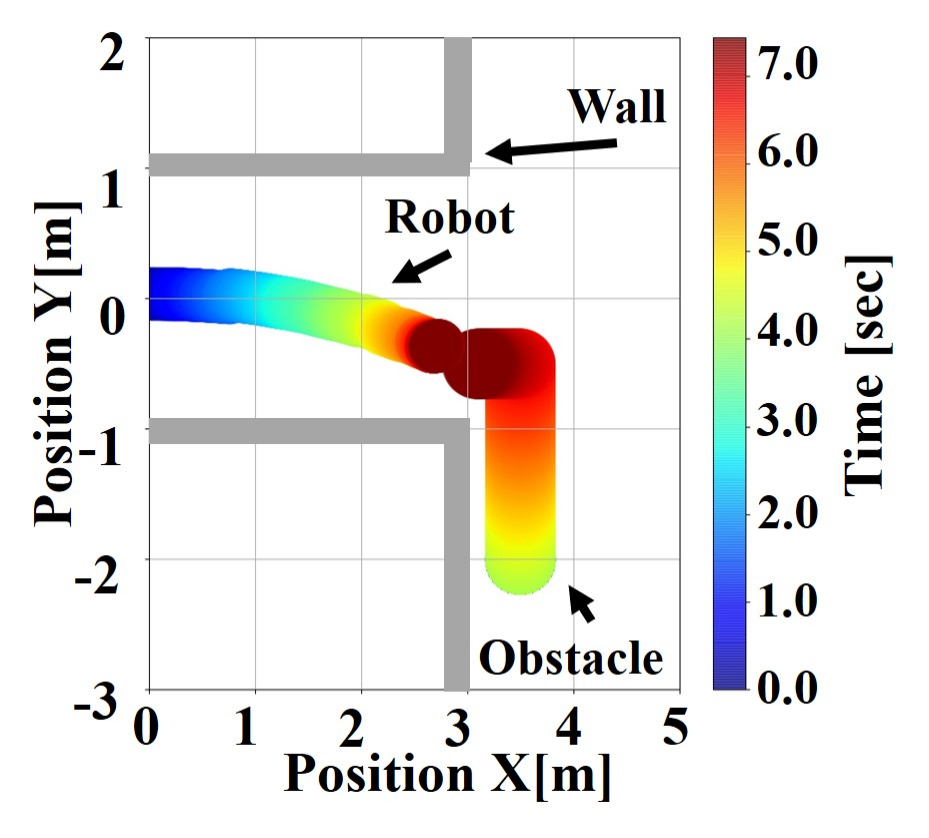}}
   \vspace{-5mm}
      {\begin{center} (a) {\it Method 1} \end{center}}
    \end{center}
  \end{minipage}
  \begin{minipage}{0.245\hsize}
    \begin{center}
      \scalebox{0.25}{
      \includegraphics{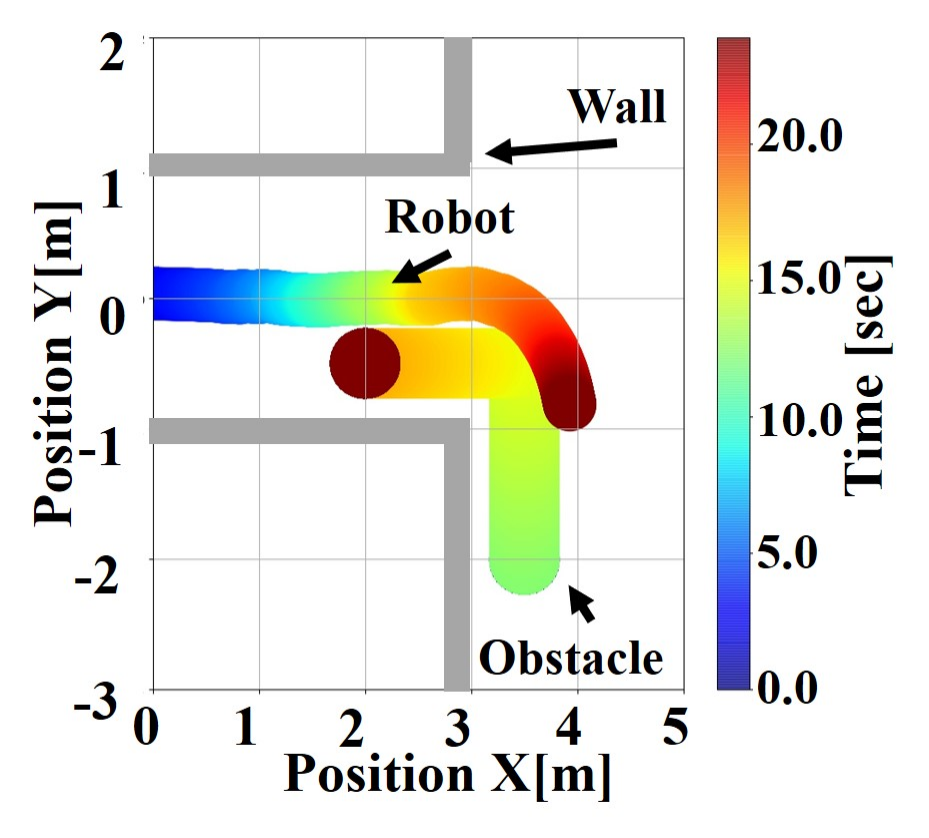}}
   \vspace{-5mm}
      {\begin{center} (b) {\it Method 2} \end{center}}
    \end{center}
  \end{minipage}
  \begin{minipage}{0.245\hsize}
    \begin{center}
      \scalebox{0.25}{
      \includegraphics{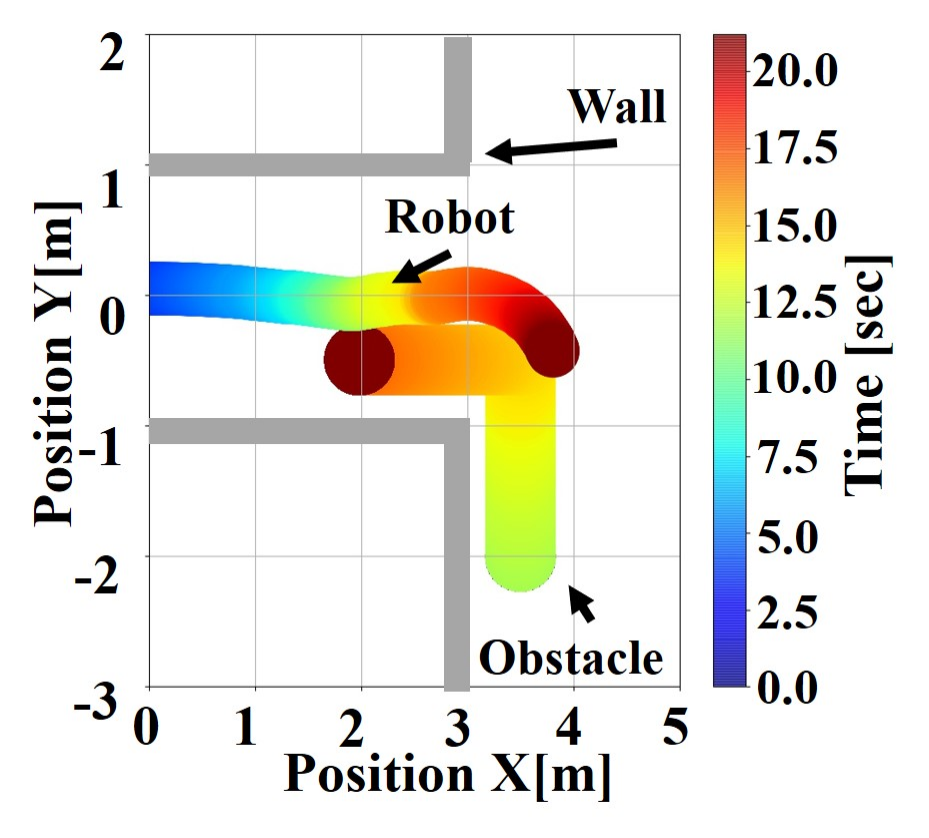}}
   \vspace{-5mm}
      {\begin{center} (c) {\it Method 3} \end{center}}
    \end{center}
  \end{minipage}
  \begin{minipage}{0.245\hsize}
    \begin{center}
      \scalebox{0.25}{
      \includegraphics{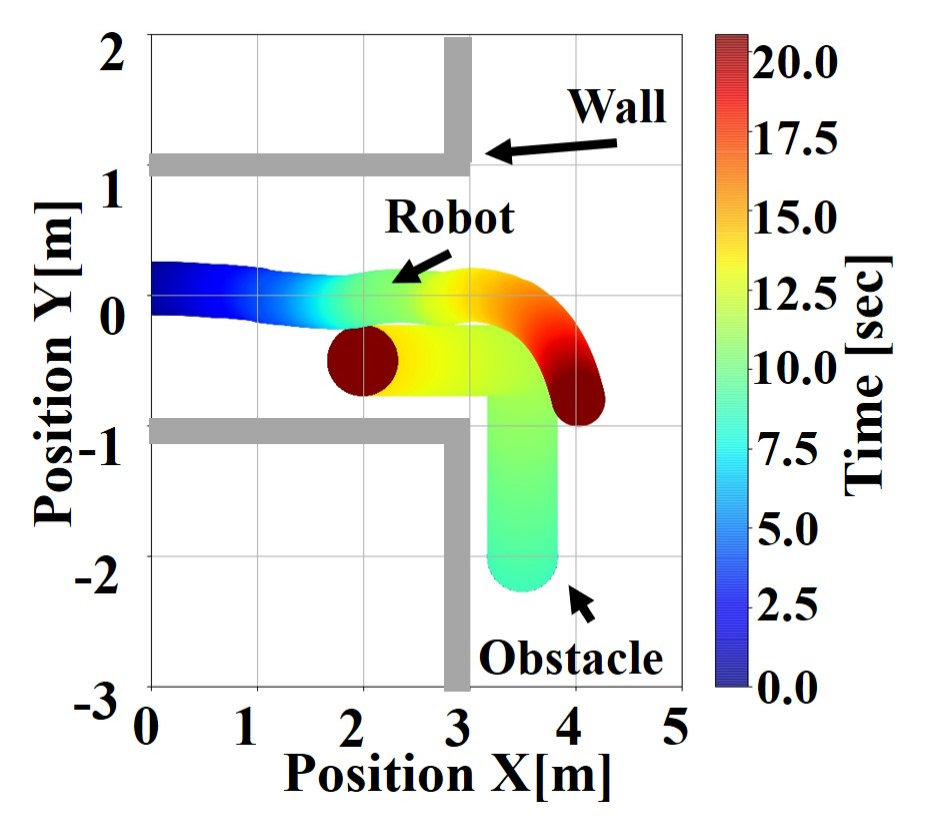}}
   \vspace{-5mm}
      {\begin{center} (d) {\it Method 4} \end{center}}
    \end{center}
  \end{minipage}
  \caption{Simulation Results (Case S2)}
  \label{fig:sim-res-rgbd2}
\end{figure*}
Table~\ref{1} shows the control parameters.
The parameters were determined by trial and error.
As shown in Fig.~\ref{fig:se}(a), the robot was equipped with the LRF and RGB-D cameras.

In this simulation, there are 2 cases; Case S1 and Case S2.
As shown in Fig.~\ref{fig:se}(b)(c), the dynamic obstacle assumed as the human was placed at the position that cannot be recognized by the robot.
When the robot crosses the green line, the dynamic obstacle moves at the velocity of 4.0 [km/h] on the orange arrow, which is assumed as the walking velocity of the human.
The robot moves by using DWA with the maximum velocity of 2.0 [km/h].

\subsubsection{Simulation Method}
Table~\ref{tab_method} shows simulation methods.
We treated the conventional methods as {\it Method 1} and {\it Method 2}, and the proposed methods as {\it Method 3} and {\it Method 4}.
Environmental information is obtained from LRF in {\it Method 1} and {\it Method 2}.
{\it Method 3} and {\it Method 4} acquire environmental information from RGB-D cameras and LRF.
Simulations were performed in Case S1 and Case S2 using the conventional and proposed methods.
\subsection{Simulation Results}

\subsubsection{Case S1}
Fig.~\ref{fig:sim-res-rgbd1} shows the simulation results in Case S1.
From Fig.~\ref{fig:sim-res-rgbd1}(a), the robot collided with the obstacle because the blind spots area was not considered in {\it Method 1}.
In {\it Method 2} - {\it Method 4}, Fig.~\ref{fig:sim-res-rgbd1}(b)-(d) show that the robot avoided the collision with the obstacle because the blind spots area was taken into account.
As shown in Table~\ref{tab2}, the goal time of the conventional method ({\it Method 2}) is 25.3[sec] and the proposed method ({\it Method 4}) is 18.3[sec].
The goal arrival time of the proposed method ({\it Method 4}) was improved by 27.7$\%$ compared with the conventional method ({\it Method 2}).

\subsubsection{Case S2}
Fig.~\ref{fig:sim-res-rgbd2} shows the simulation results in Case S2.
From Fig.~\ref{fig:sim-res-rgbd2}(a), the robot collided with the obstacle because the blind spots area was not considered in {\it Method 1}.
In {\it Method 2} and {\it Method 4}, Fig.~\ref{fig:sim-res-rgbd2}(b)-(d) show that the robot avoided the collision with the obstacle because the blind spots area was taken into account.
As shown in Table~\ref{tab2}, the goal time of the conventional method ({\it Method 2}) is 23.8[sec] and the proposed method ({\it Method 4}) is 20.5[sec].
The goal arrival time of the proposed method ({\it Method 4}) was improved by 13.9$\%$ compared with the conventional method ({\it Method 2}).

\subsubsection{Discussion}
\begin{figure}[t]
  \begin{minipage}{0.49\hsize}
    \begin{center}
      \scalebox{0.18}{
        \includegraphics{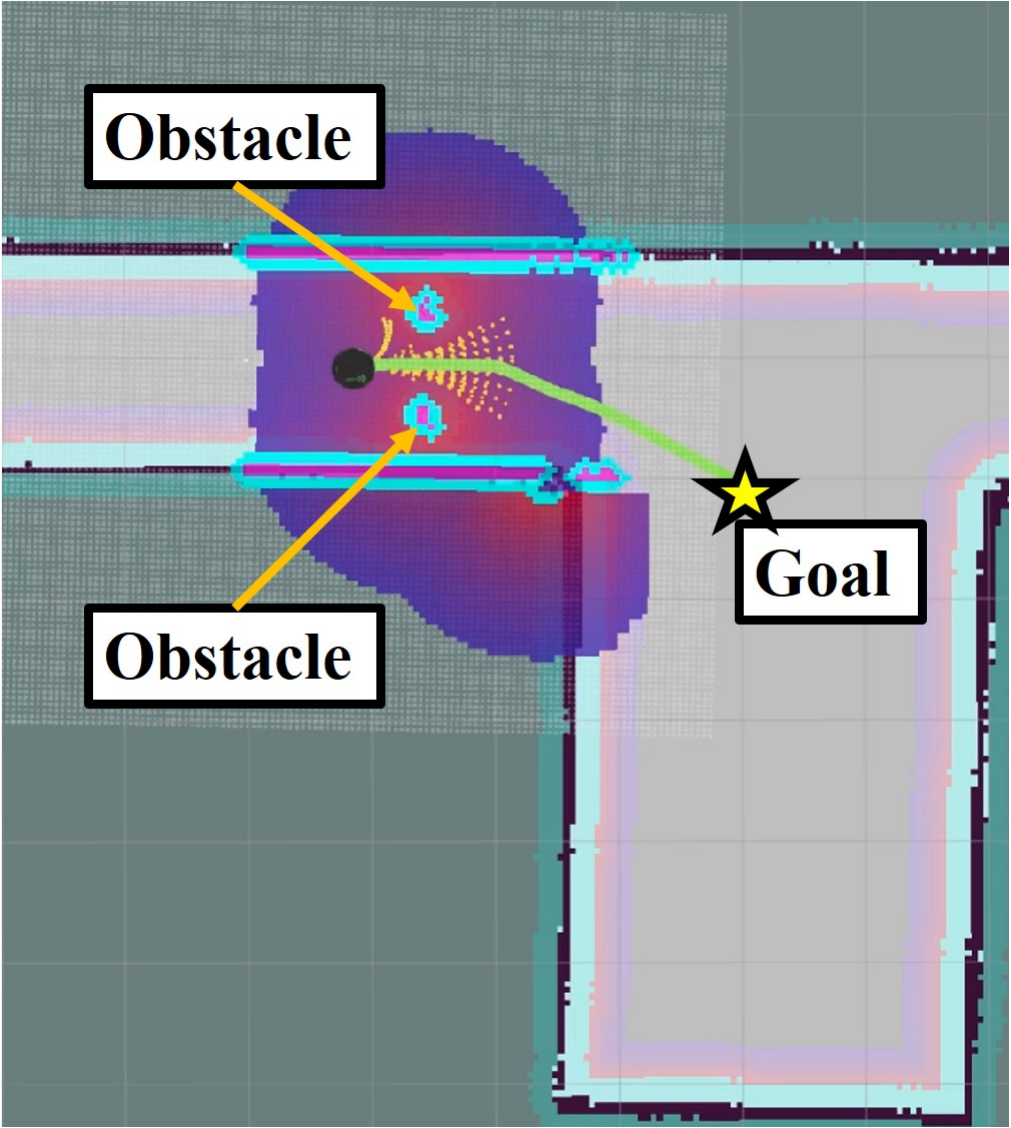}}
      {\begin{center} (a) Method 2\end{center}}
    \end{center}
  \end{minipage}
  \begin{minipage}{0.49\hsize}
    \begin{center}
      \scalebox{0.18}{
        \includegraphics{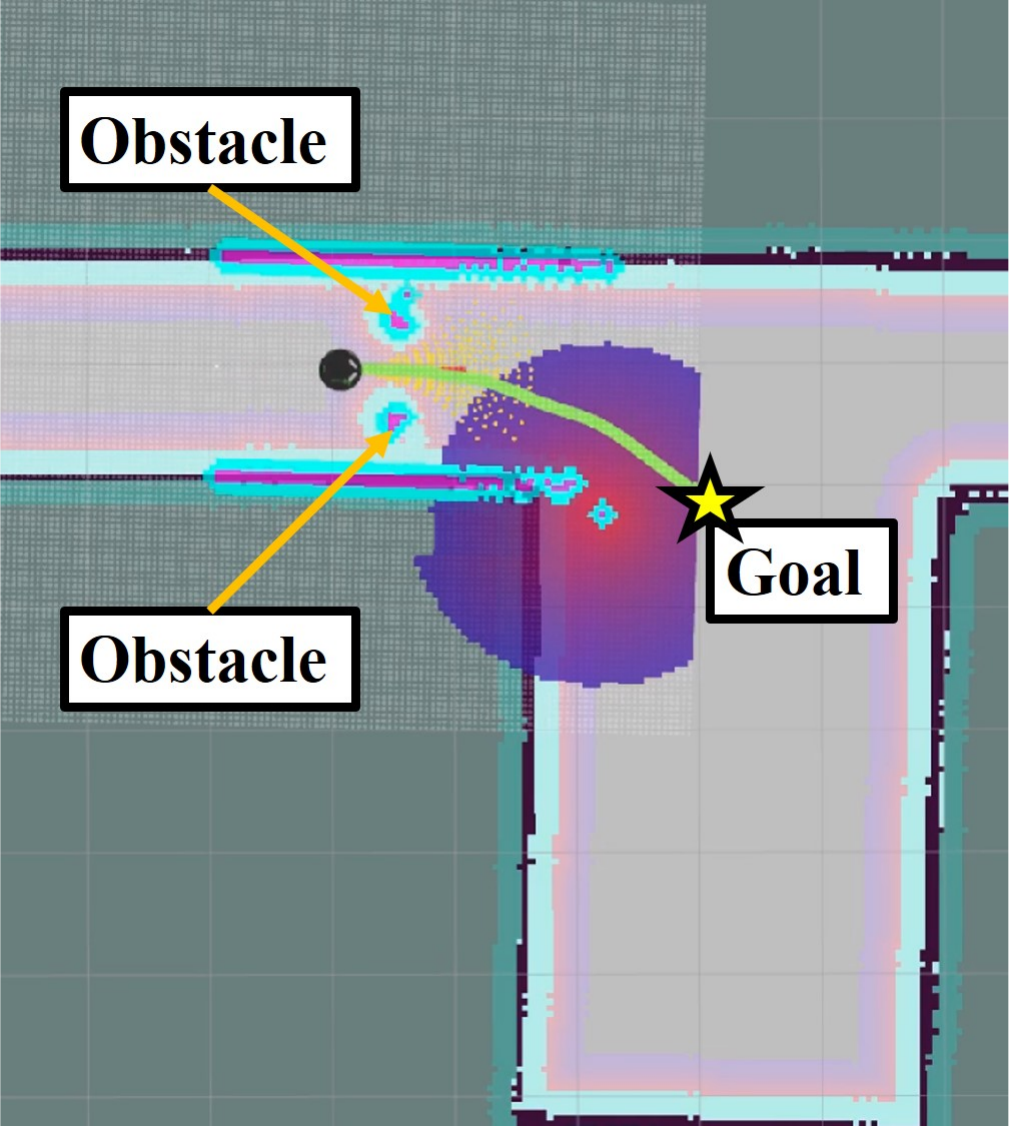}}
      {\begin{center} (b) Method 4\end{center}}
    \end{center}
  \end{minipage}
  \caption{Comparison between Method 2 and Method 4}
  \label{fig:sr}
\end{figure}

There were two reasons why the proposed method had a faster arrival time than the conventional method.
Firstly, as shown in Fig.~\ref{fig:sr}, the conventional method generated the dangerous area only by the LRF. 
Thus, the conventional method ({\it Method 2}) redundantly generated the dangerous area even for small obstacles.
In the proposed method ({\it Method 4}), the dangerous area was estimated by RGB-D cameras, so that small obstacles were excluded.
Therefore, the proposed method prevented the redundant generation of dangerous regions.
Secondly, the proposed method added the velocity term in (10), which made the arrival time shorter than the conventional method.

The effectiveness of the proposed method was confirmed by the simulation results of Case S1 and Case S2.

\section{Experiment}
\subsection{Experiment Setup}
\begin{figure*}[t]
  \begin{minipage}{0.32\hsize}
    \begin{center}
      \scalebox{0.32}{
          \includegraphics{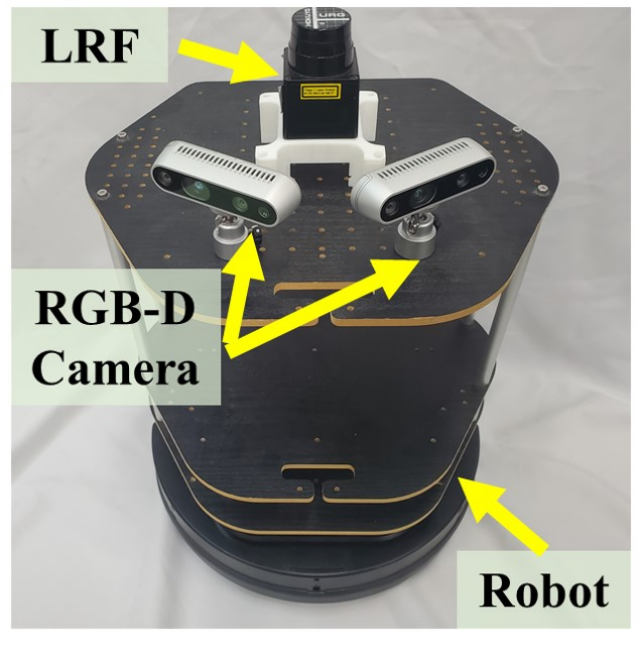}}
      \vspace{-1mm}
      {\begin{center} (a) Robot\end{center}}
    \end{center}
  \end{minipage}
  \begin{minipage}{0.32\hsize}
    \begin{center}
      \scalebox{0.35}{
         \includegraphics{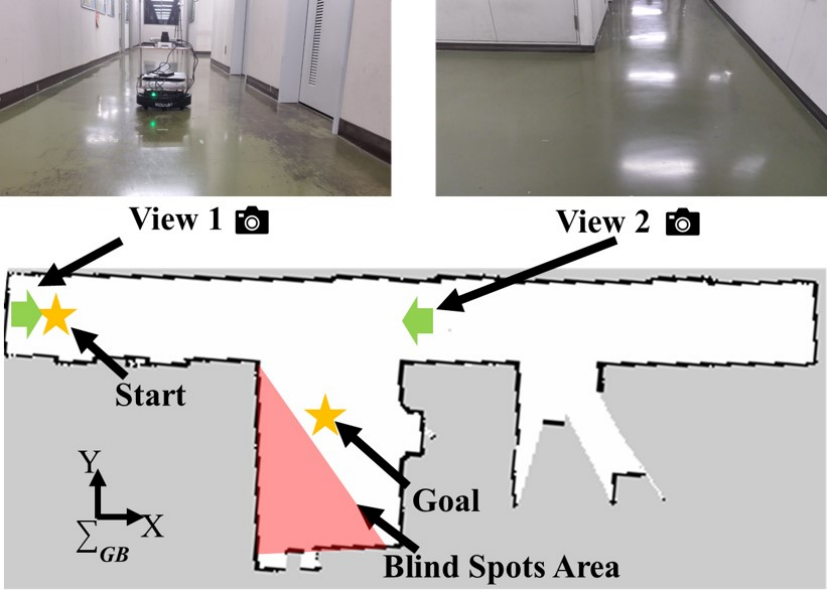}}
      \vspace{-1mm}
      {\begin{center} (b) Environment (Case E1)\end{center}}
    \end{center}
  \end{minipage}
  \begin{minipage}{0.32\hsize}
    \begin{center}
      \scalebox{0.35}{
         \includegraphics{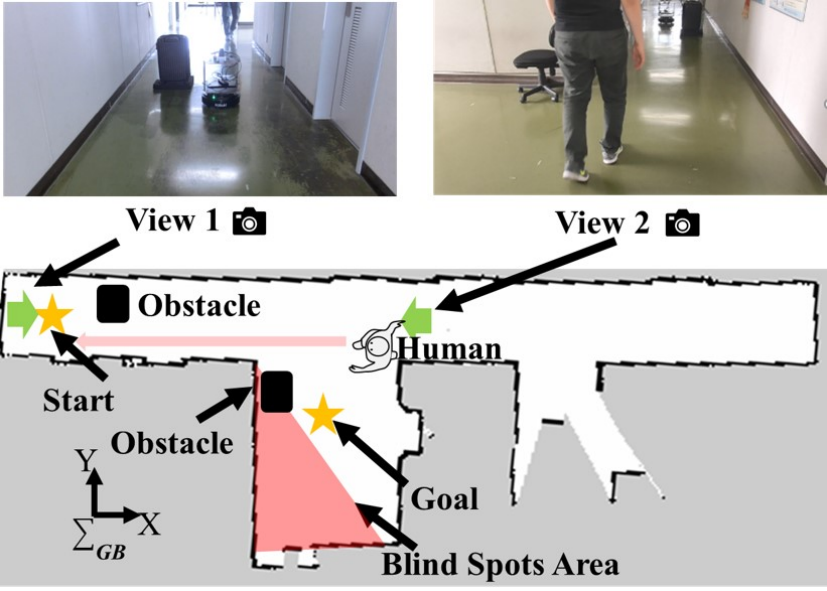}}
      \vspace{-1mm}
      {\begin{center} (c) Environment (Case E2)\end{center}}
    \end{center}
  \end{minipage}
  \caption{Experiments Setup}
  \label{fig:ex}
\end{figure*}

As shown in Fig.~\ref{fig:ex} (a), the robot was equipped with the LRF (URG-04LX-UG01) and RGB-D cameras (Intel RealSense D435i).
The proposed system was implemented by ROS.
As shown in Fig.~\mbox{\ref{fig:ex}}(b)(c), there are 2 cases; Case E1 and Case E2 in this experiment.
In Case E1, we conducted experiments in an environment with no obstructions but with the existence of blind spots, to confirm whether the proposed method operates on the real robot.
In Case E2, we carried out experiments in an environment where there was one obstacle in the blind spot area, one outside of it, and a pedestrian was present.
As shown in Table~\ref{1}, the same parameters as in the simulation were set for the experiment.

\subsection{Experiment Results}

\begin{figure}[t]
      \vspace{-4mm}
  \begin{minipage}{0.49\hsize}
    \begin{center}
      \scalebox{0.18}{
         \includegraphics{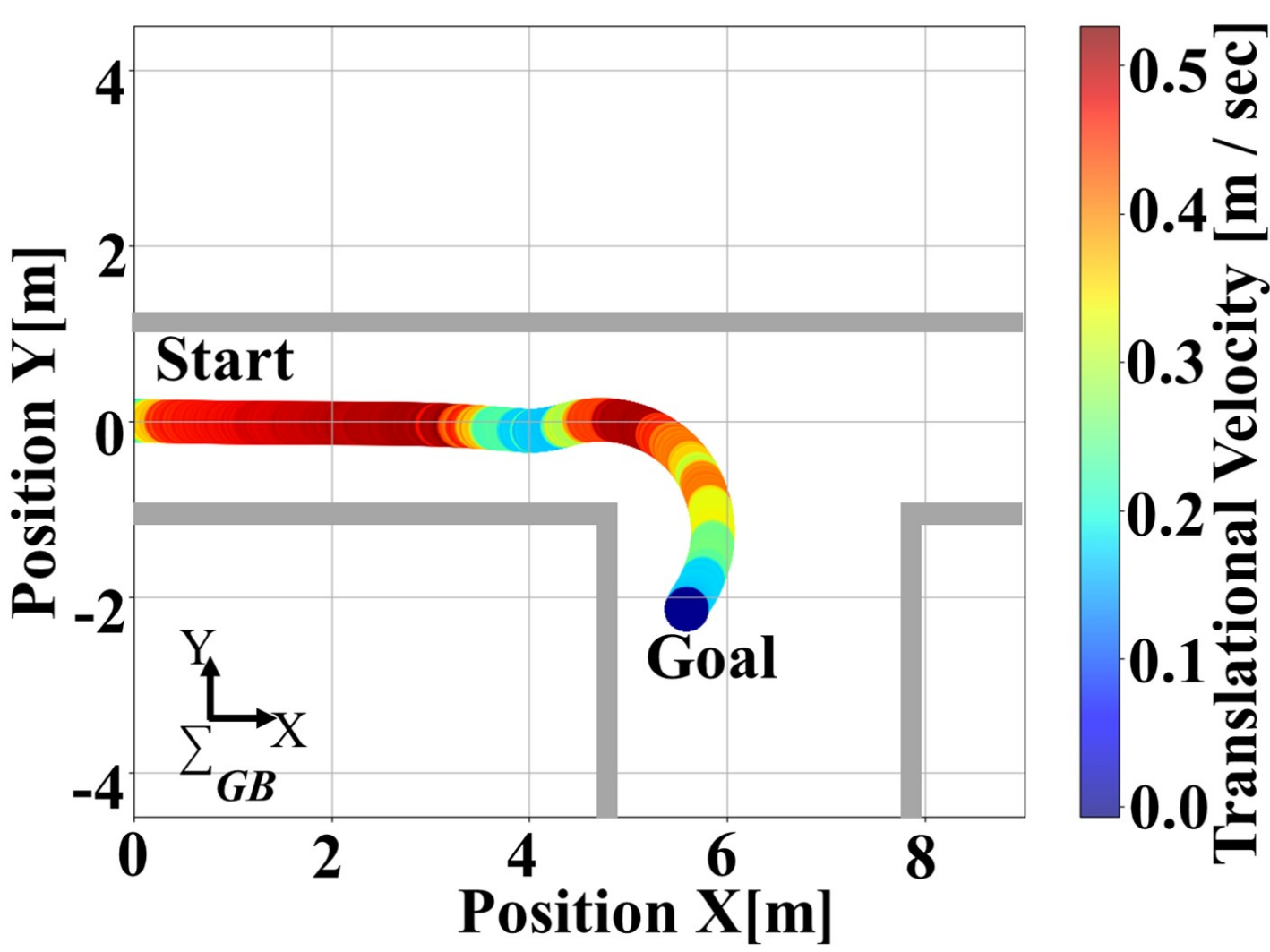}}
      \vspace{-2mm}
      {\begin{center} (a) Case E1\end{center}}
    \end{center}
  \end{minipage}
  \begin{minipage}{0.49\hsize}
    \begin{center}
      \scalebox{0.18}{
         \includegraphics{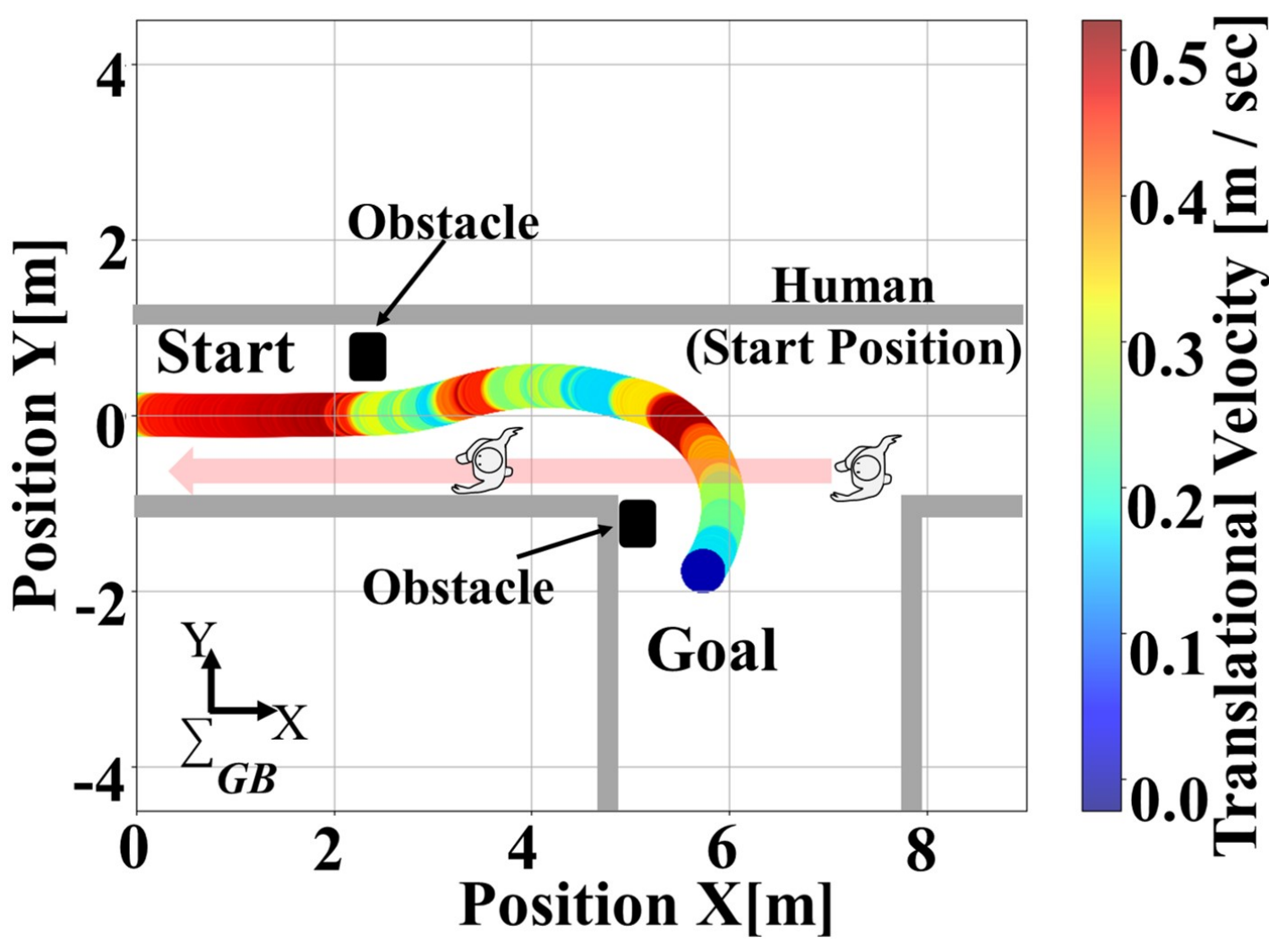}}
      \vspace{-2mm}
      {\begin{center} (b) Case E2\end{center}}
    \end{center}
  \end{minipage}
  \caption{Trajectory Results}
  \label{fig:ex-r}
\end{figure}

\begin{figure*}[!t]
  \begin{minipage}{0.245\hsize}
    \begin{center}
      \scalebox{0.44}{
        \includegraphics{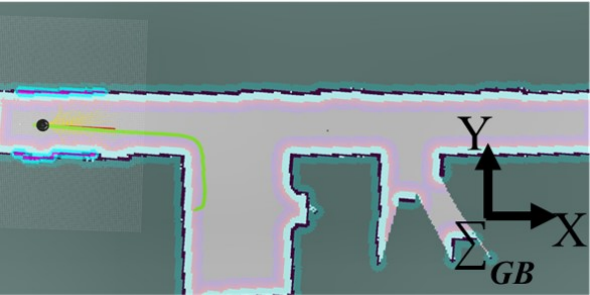}}
      {\begin{center} (a) Situation 1 (2 [sec])\end{center}}
    \end{center}
  \end{minipage}
  \begin{minipage}{0.245\hsize}
    \begin{center}
      \scalebox{0.44}{
        \includegraphics{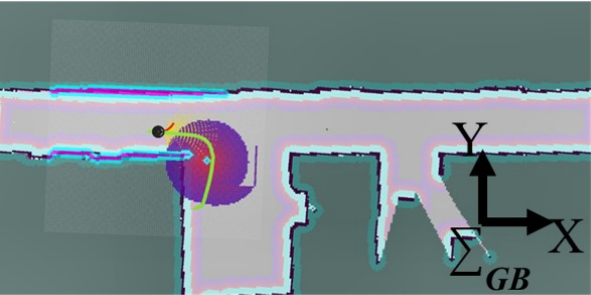}}
      {\begin{center} (b) Situation 2 (8 [sec])\end{center}}
    \end{center}
  \end{minipage}
  \begin{minipage}{0.245\hsize}
    \begin{center}
      \scalebox{0.44}{
         \includegraphics{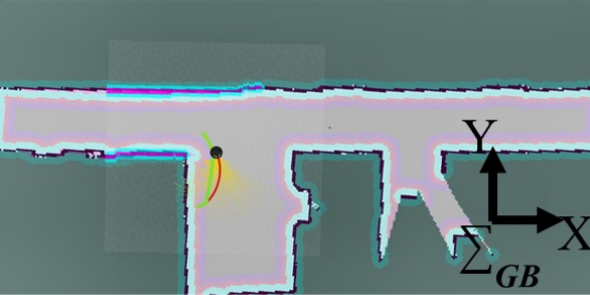}}
      {\begin{center} (c) Situation 3 (16 [sec])\end{center}}
    \end{center}
  \end{minipage}
  \begin{minipage}{0.245\hsize}
    \begin{center}
      \scalebox{0.44}{
         \includegraphics{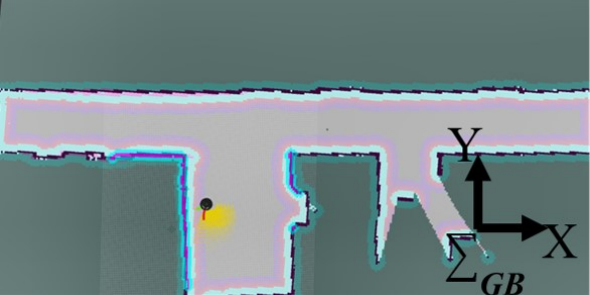}}
      {\begin{center} (d) Situation 4 (20 [sec]) \end{center}}
    \end{center}
  \end{minipage}
  \caption{Case E1 Results (Cost map)}
  \label{fig:er_costmap1}
\end{figure*}

\begin{figure*}[!t]
  \begin{minipage}{0.245\hsize}
    \begin{center}
      \scalebox{0.44}{
         \includegraphics{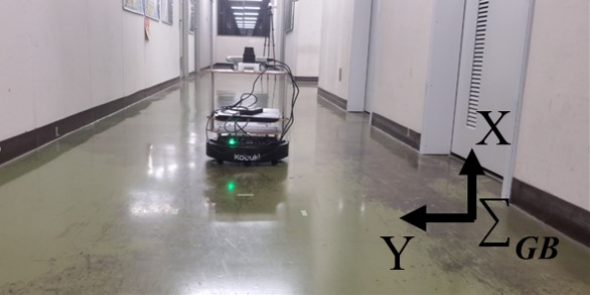}}
      {\begin{center} (a) View 1 (2 [sec])\end{center}}
    \end{center}
  \end{minipage}
  \begin{minipage}{0.245\hsize}
    \begin{center}
      \scalebox{0.44}{
         \includegraphics{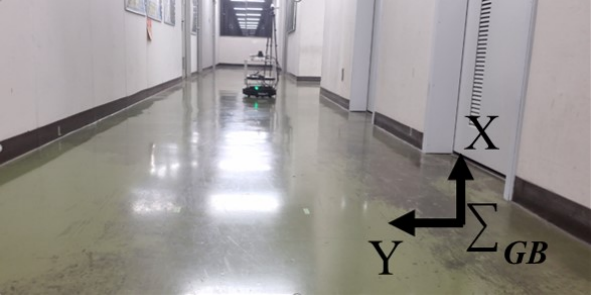}}
      {\begin{center} (b) View 1 (8 [sec])\end{center}}
    \end{center}
  \end{minipage}
  \begin{minipage}{0.245\hsize}
    \begin{center}
      \scalebox{0.44}{
         \includegraphics{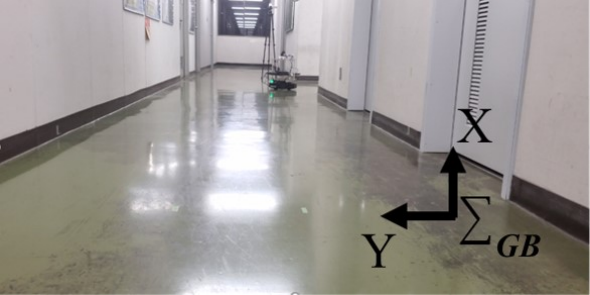}}
      {\begin{center} (c) View 1 (16 [sec])\end{center}}
    \end{center}
  \end{minipage}
  \begin{minipage}{0.245\hsize}
    \begin{center}
      \scalebox{0.44}{
         \includegraphics{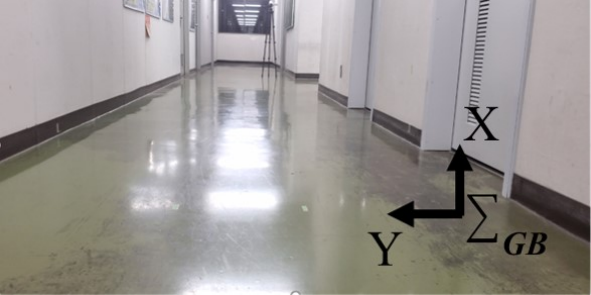}}
      {\begin{center} (d) View 1 (20 [sec]) \end{center}}
    \end{center}
  \end{minipage}
  \caption{Case E1 Results (View 1)}
  \label{fig:er_v11}
\end{figure*}

\begin{figure*}[!t]
  \begin{minipage}{0.245\hsize}
    \begin{center}
      \scalebox{0.44}{
         \includegraphics{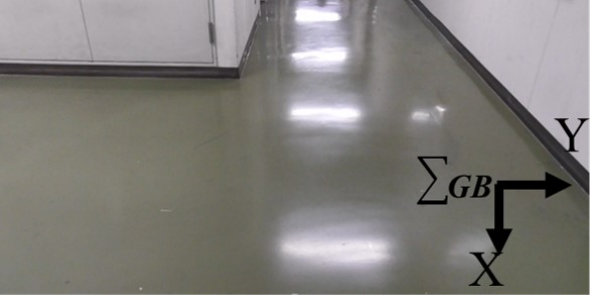}}
      {\begin{center} (a) View 2 (2 [sec]) \end{center}}
    \end{center}
  \end{minipage}
  \begin{minipage}{0.245\hsize}
    \begin{center}
      \scalebox{0.44}{
         \includegraphics{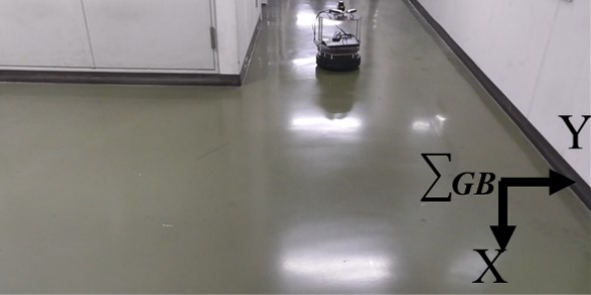}}
      {\begin{center} (b) View 2 (8 [sec])\end{center}}
    \end{center}
  \end{minipage}
  \begin{minipage}{0.245\hsize}
    \begin{center}
      \scalebox{0.44}{
         \includegraphics{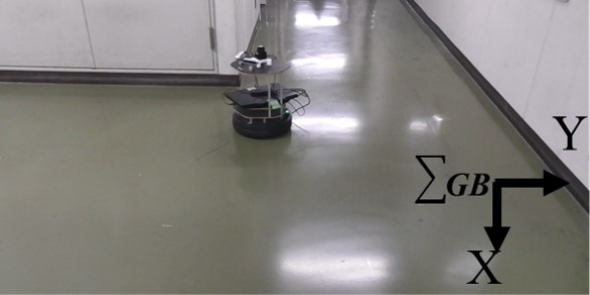}}
      {\begin{center} (c) View 2 (16 [sec])\end{center}}
    \end{center}
  \end{minipage}
  \begin{minipage}{0.245\hsize}
    \begin{center}
      \scalebox{0.44}{
         \includegraphics{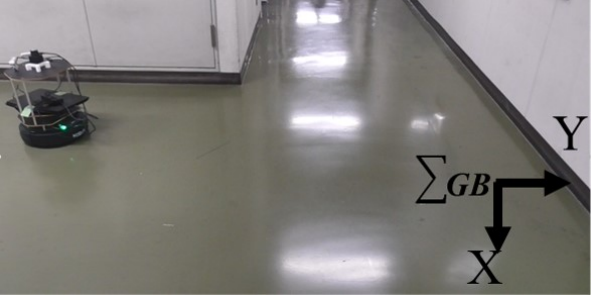}}
      {\begin{center} (d) View 2 (20 [sec])\end{center}}
    \end{center}
  \end{minipage}
  \caption{Case E1 Results (View 2)}
  \label{fig:er_v21}
\end{figure*}

\begin{figure*}[!t]
  \begin{minipage}{0.245\hsize}
    \begin{center}
      \scalebox{0.44}{
        \includegraphics{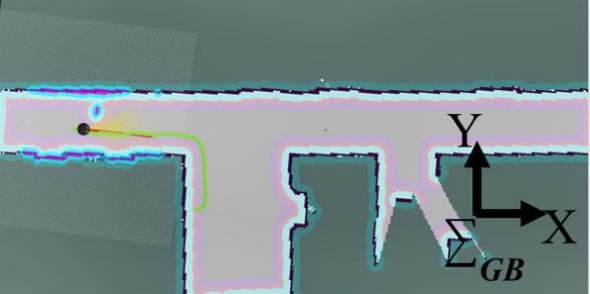}}
      {\begin{center} (a) Situation 1 (4 [sec])\end{center}}
    \end{center}
  \end{minipage}
  \begin{minipage}{0.245\hsize}
    \begin{center}
      \scalebox{0.44}{
        \includegraphics{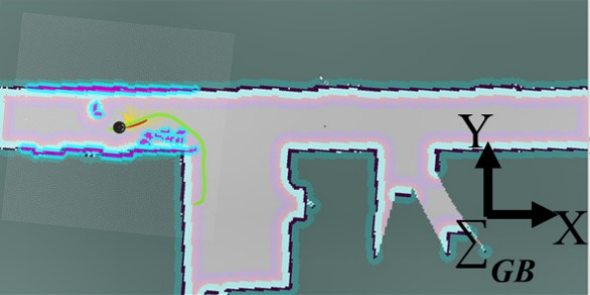}}
      {\begin{center} (b) Situation 2 (7 [sec])\end{center}}
    \end{center}
  \end{minipage}
  \begin{minipage}{0.245\hsize}
    \begin{center}
      \scalebox{0.44}{
         \includegraphics{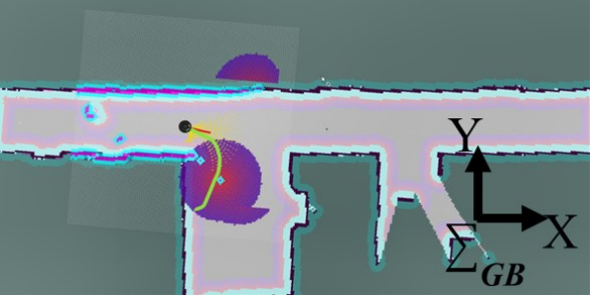}}
      {\begin{center} (c) Situation 3 (14 [sec])\end{center}}
    \end{center}
  \end{minipage}
  \begin{minipage}{0.245\hsize}
    \begin{center}
      \scalebox{0.44}{
         \includegraphics{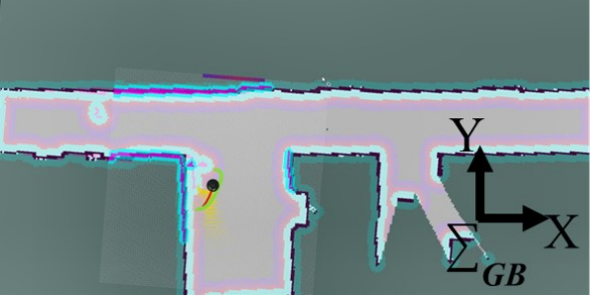}}
      {\begin{center} (d) Situation 4 (22 [sec]) \end{center}}
    \end{center}
  \end{minipage}
  \caption{Case E2 Results (Cost map)}
  \label{fig:er_costmap2}
\end{figure*}

\begin{figure*}[!t]
  \begin{minipage}{0.245\hsize}
    \begin{center}
      \scalebox{0.44}{
         \includegraphics{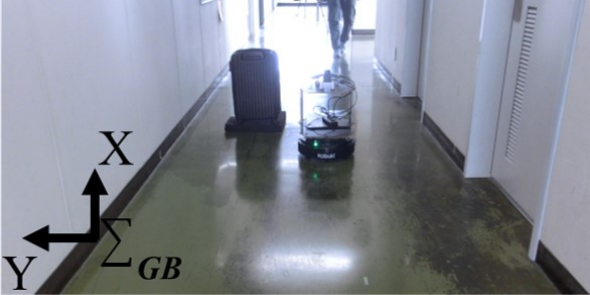}}
      {\begin{center} (a) View 1 (4 [sec])\end{center}}
    \end{center}
  \end{minipage}
  \begin{minipage}{0.245\hsize}
    \begin{center}
      \scalebox{0.44}{
         \includegraphics{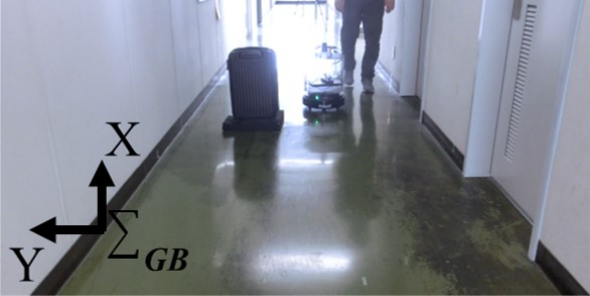}}
      {\begin{center} (b) View 1 (7 [sec])\end{center}}
    \end{center}
  \end{minipage}
  \begin{minipage}{0.245\hsize}
    \begin{center}
      \scalebox{0.44}{
         \includegraphics{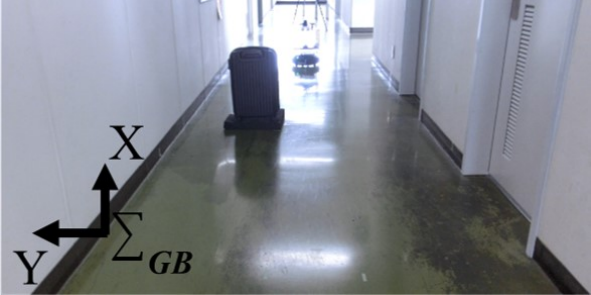}}
      {\begin{center} (c) View 1 (14 [sec])\end{center}}
    \end{center}
  \end{minipage}
  \begin{minipage}{0.245\hsize}
    \begin{center}
      \scalebox{0.44}{
         \includegraphics{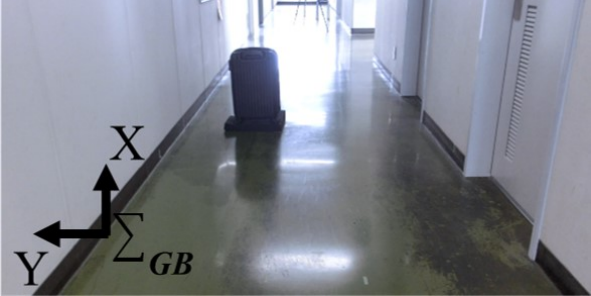}}
      {\begin{center} (d) View 1 (22 [sec]) \end{center}}
    \end{center}
  \end{minipage}
  \caption{Case E2 Results (View 1)}
  \label{fig:er_v12}
\end{figure*}

\begin{figure*}[!t]
  \begin{minipage}{0.245\hsize}
    \begin{center}
      \scalebox{0.44}{
         \includegraphics{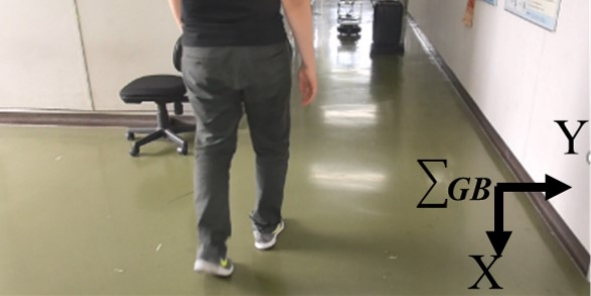}}
      {\begin{center} (a) View 2 (4 [sec]) \end{center}}
    \end{center}
  \end{minipage}
  \begin{minipage}{0.245\hsize}
    \begin{center}
      \scalebox{0.44}{
         \includegraphics{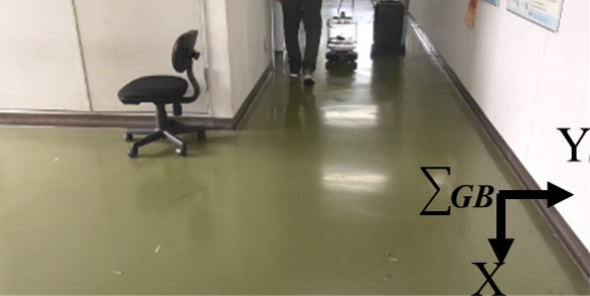}}
      {\begin{center} (b) View 2 (7 [sec])\end{center}}
    \end{center}
  \end{minipage}
  \begin{minipage}{0.245\hsize}
    \begin{center}
      \scalebox{0.44}{
         \includegraphics{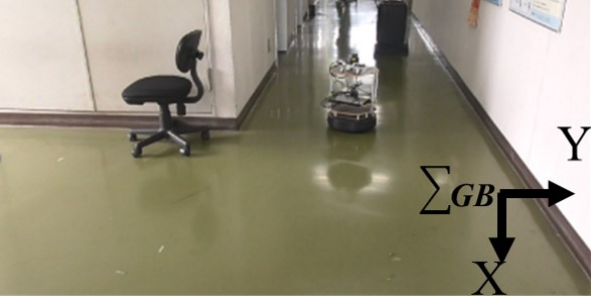}}
      {\begin{center} (c) View 2 (14 [sec])\end{center}}
    \end{center}
  \end{minipage}
  \begin{minipage}{0.245\hsize}
    \begin{center}
      \scalebox{0.44}{
         \includegraphics{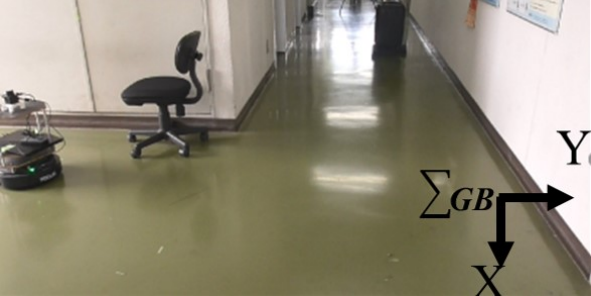}}
      {\begin{center} (d) View 2 (22 [sec])\end{center}}
    \end{center}
  \end{minipage}
  \caption{Case E2 Results (View 2)}
  \label{fig:er_v22}
\end{figure*}

Fig.\mbox{\ref{fig:ex-r}} shows the experimental trajectory results, with the color bar indicating velocity from minimum to maximum. The cost map results and snapshots from two views of the experiment are shown in Fig.\mbox{\ref{fig:er_costmap1}} -\mbox{\ref{fig:er_v22}}.

In Case E1, as depicted in Fig.\mbox{\ref{fig:ex-r}}-\mbox{\ref{fig:er_v21}}, the robot arrived at the goal using our method.
Fig.\mbox{\ref{fig:er_costmap1}}-\mbox{\ref{fig:er_v21}}(a) shows the path generated by the global path planning method.
As in Fig.\mbox{\ref{fig:er_costmap1}}-\mbox{\ref{fig:er_v21}}(b), BSL produced the blind spot cost, enabling the robot to avoid this area and slow down, as seen in Fig.~\mbox{\ref{fig:ex-r}}(a), \mbox{\ref{fig:er_costmap1}}.
The blind spot area is eliminated in Fig.\mbox{\ref{fig:er_costmap1}}-\mbox{\ref{fig:er_v21}}(c) and a local path is chosen to follow the global plan. The robot reached its goal as shown in Fig.\mbox{\ref{fig:er_costmap1}}-\mbox{\ref{fig:er_v21}}(d).

In Case E2, Fig.\mbox{\ref{fig:ex-r}},\mbox{\ref{fig:er_costmap2}}-\mbox{\ref{fig:er_v22}} shows that the robot reached the goal via our method. The global path planning method generated a path from start to goal, as seen in Fig.\mbox{\ref{fig:er_costmap2}}-\mbox{\ref{fig:er_v22}}(a), with the robot recognizing and avoiding an obstacle outside its blind spot.
The robot also detected a pedestrian and executed collision avoidance, as shown in Fig.\mbox{\ref{fig:er_costmap2}}-\mbox{\ref{fig:er_v22}}(b).
As shown in Fig.~\mbox{\ref{fig:er_costmap2}}-\mbox{\ref{fig:er_v22}}(c), 
BSL generated the blind spots cost. Thus, the robot avoided the blind spots area and reduce the velocity from Fig.~\mbox{\ref{fig:ex-r}}(b), \mbox{\ref{fig:er_costmap2}}.
As shown in Fig.~\mbox{\ref{fig:er_costmap2}}-\mbox{\ref{fig:er_v22}}(d), the blind spot area was eliminated and the local path was selected to follow the global path plan. 
The robot arrived at the goal position.

The proposed method successfully considered the blind spot area in real environments. The experimental results confirmed the effectiveness of our method.

\section{Conclusion}
This paper proposed the navigation method considering blind spots based on the robot operating system (ROS) navigation stack and blind spots layer for a wheeled mobile robot.
Blind spots occur when the robot approaches corners or obstacles.
If the human or object moves toward the robot from blind spots, a collision may occur.
For collision avoidance, this paper describes local path planning considering blind spots.
Blind spots are estimated from the environmental information measured by RGB-D cameras.
In the proposed method, path planning considering blind spots is achieved by the cost map ``BSL'' and ``DWA'' which is local path planning with an improved cost function.
The effectiveness of the proposed method was further demonstrated through simulations.

In future works, we will work to evaluate our method as follows.
\begin{itemize}
  \item {\it Parameter Design of BSL}\\
        The number of parameters was increased by considering BSL.
        The parameter design method should be clarified and improved. We will adopt a machine learning method to determine BSL parameters\mbox{\cite{drl}}.

  \item {\it BSL with Various Path Planning}\\
        We consider combining BSL with any path planning method that can handle cost map and explore alternative approaches.

  \item {\it Various Environments, Sensors, and Robots}\\
        We evaluated BSL with the robot with RGB-D cameras and environments.
        We will evaluate BSL for various robots, sensors, and environments.
        Especially, we would like to integrate RGB-D and LiDAR\mbox{\cite{rgbd-1, LD-1}}.

  \item {\it ROS 2}\\
        We have implemented BSL using the ROS Navigation Stack. We will implement it with ROS 2\mbox{\cite{nav2}}.

\end{itemize}

\section*{Acknowledgments}
This work was supported in part by the Kansai Research Foundation for Technology Promotion.

\vfill


\begin{thebibliography}{1}
\bibliographystyle{IEEEtran}
  \bibitem{kobayashi2022bsl}
  M. Kobayashi and N. Motoi, ``Path Planning Method Considering Blind Spots Based on ROS Navigation Stack and Dynamic Window Approach for Wheeled Mobile Robot,'' {\it Proceedings of International Power Electronics Conference}, pp.~274-279, 2022.

  \bibitem{Teeneti2021wheelchairs}
    C. R. Teeneti, U. Pratik, G. R. Philips, A. Azad, M. Greig, R. Zane, C. Bodine, C. Coopmans, and Z. Pantic, ``System-Level Approach to Designing a Smart Wireless Charging System for Power Wheelchairs,'' {\it IEEE Transactions on Industry Applications}, vol.~57, no.~5, pp.~5128-5144, 2021.

  \bibitem{wang2022medicalrobot}
   J. Wang, C. Yue, G. Wang, Y. Gong, H. Li, W. Yao, S. Kuang, W. Liu, J. Wang, and B. Su, ``Task Autonomous Medical Robot for Both Incision Stapling and Staples Removal,'' {\it IEEE Robotics and Automation Letters}, vol.~7, no.~2, pp.~3279-3285, 2022.

  \bibitem{liao2023mwh}
    L. Cai, Z. Liao, S. Wei, and J. Li, ``Novel Direct Yaw Moment Control of Multi-Wheel Hub Motor Driven Vehicles for Improving Mobility and Stability,'' {\it IEEE Transactions on Industry Applications}, vol.~59, no.~1, pp.~591-600, 2023.

  \bibitem{kumar2021surveyhrc}
    S. Kumar, C. Savur, and F. Sahin, ``Survey of Human–Robot Collaboration in Industrial Settings: Awareness, Intelligence, and Compliance,'' {\it IEEE Transactions on Systems, Man, and Cybernetics: Systems}, vol.~51, no.~1, pp.~280-297, 2021.
    
 \bibitem{han2022srg}
    S. Han, S. Chon, J. Kim, J. Seo, D. G. Shin, S. Park, J. T. Kim, J. Kim, M. Jin, and J. Cho., ``Snake Robot Gripper Module for Search and Rescue in Narrow Spaces,'' {\it IEEE Robotics and Automation Letters}, vol.~7, no.~2, pp.~1667-1673, 2022.

 \bibitem{seeja2022snake}
    G. Seeja, A. Selvakumar Arockia Doss, and V. B. Hency, ``A Survey on Snake Robot Locomotion,'' {\it IEEE Access}, vol.~10, pp.~112100-112116, 2022.
        
  \bibitem{abegaz2022food}
    B. W. Abegaz, ``A Parallelized Self-Driving Vehicle Controller Using Unsupervised Machine Learning,'' {\it IEEE Transactions on Industry Applications}, vol.~58, no.~4, pp.~5148-5156, 2022.

  \bibitem{saito2021robot}
    N. Saito, T. Ogata, S. Funabashi, H. Mori, and S. Sugano, ``How to Select and Use Tools? : Active Perception of Target Objects Using Multimodal Deep Learning,'' {\it IEEE Robotics and Automation Letters}, vol.~6, no.~2, pp.~2517-2524, 2021.
    
  \bibitem{nagpal2018mani}
    N. Nagpal, V. Agarwal, and B. Bhushan, ``A Real-Time State-Observer-Based Controller for a Stochastic Robotic Manipulator,'' {\it IEEE Transactions on Industry Applications}, vol.~54, no.~2, pp.~1806-1822, 2018.

  \bibitem{aziz2019mani}
  M. A. S. Aziz, S. Yahya, H. A. F. Almurib, Y. A. Abakr, M. Moghavvemi, Z. Madibekov, A. S. A. Elsayed, and M. O. M. AbdulRazic, ``Torque Minimized Design of a Light Weight 3 DoF Planar Manipulator,'' {\it IEEE Transactions on Industry Applications}, vol.~55, no.~3, pp.~3207-3214, 2019.

  \bibitem{martin2021mm1}
    J. Martin, A. Ansuategi, I. Maurtua, A. Gutierrez, D. Obregón, O. Casquero, and M. Marcos, ``A Generic ROS-Based Control Architecture for Pest Inspection and Treatment in Greenhouses Using a Mobile Manipulator,'' {\it IEEE Access}, vol.~9, pp.~94981-94995, 2021.
    
  \bibitem{selvaggio2021hri}
    M. Selvaggio, M. Cognetti, S. Nikolaidis, S. Ivaldi, and B. Siciliano, ``Autonomy in Physical Human-Robot Interaction: A Brief Survey,'' {\it IEEE Robotics and Automation Letters}, vol.~6, no.~4, pp.~7989-7996, 2021.
    
  \bibitem{zhang2023service}
    Y. Zhang, G. Tian, X. Shao, M. Zhang, and S. Liu, ``Semantic Grounding for Long-Term Autonomy of Mobile Robots Toward Dynamic Object Search in Home Environments,'' {\it IEEE Transactions on Industrial Electronics}, vol.~70, no.~2, pp.~1655-1665, 2023.
        
  \bibitem{bae2022local}
    J. Bae and D. -H. Lee, ``PTP Tracking Scheme for Indoor Surveillance Vehicle by Dual BLACM With Hall Sensor,'' {\it IEEE Transactions on Industry Applications}, vol.~58, no.~4, pp.~5238-5247, 2022.

  \bibitem{TIA-mapping}
    Y. Zheng, S. Chen, and H. Cheng, ``Real-Time Cloud Visual Simultaneous Localization and Mapping for Indoor Service Robots,'' {\it IEEE Access}, vol.~8, pp.~16816-16829, 2020.

  \bibitem{TIA-perception}
    M. B. Alatise and G. P. Hancke, ``A Review on Challenges of Autonomous Mobile Robot and Sensor Fusion Methods,'' {\it IEEE Access}, vol.~8, pp.~39830-39846, 2020.

  \bibitem{TIA-pathplanning}
    C. Ji, Y. Liu, L. Lyu, X. Li, C. Liu, Y. Peng, and Y. Xiang, ``A Personalized Fast-Charging Navigation Strategy Based on Mutual Effect of Dynamic Queuing,'' {\it IEEE Transactions on Industry Applications}, vol.~56, no.~5, pp.~5729-5740, 2020.

  \bibitem{park2014robot}
    C. Park, S. Lee, G. -H. Cho, S. -Y. Choi, and C. T. Rim, ``Two-Dimensional Inductive Power Transfer System for Mobile Robots Using Evenly Displaced Multiple Pickups,'' {\it IEEE Transactions on Industry Applications}, vol.~50, no.~1, pp.~558-565, 2014.

  \bibitem{kurita2011motion}
    K. Kurita and S. Ueta, ``A New Motion Control Method for Bipedal Robot Based on Noncontact and Nonattached Human Motion Sensing Technique,'' {\it IEEE Transactions on Industry Applications}, vol.~47, no.~2, pp.~1022-1027, 2011.

  \bibitem{kobayashi2022dwv}
    M. Kobayashi and N. Motoi, ``Local Path Planning: Dynamic Window Approach With Virtual Manipulators Considering Dynamic Obstacles,'' {\it IEEE Access}, vol.~10, pp.~17018-17029, 2022.

  \bibitem{mondal2020con}
    R. Mondal and J. Dey, ``Performance Analysis and Implementation of Fractional Order 2-DOF Control on Cart–Inverted Pendulum System,'' {\it IEEE Transactions on Industry Applications}, vol.~56, no.~6, pp.~7055-7066, 2020.

  \bibitem{schlegel2021bsl}
    K. Schlegel, P. Weissig, and P. Protzel, ``A blind-spot-aware optimization-based planner for safe robot navigation,'' {\it Proceedings of European Conference on Mobile Robots}, pp.~1-8, 2021.

  \bibitem{zhu2020bsl}
    L. Zhu, M. Menon, M. Santillo, and G. Linkowski, ``Occlusion Handling for Industrial Robots,'' {\it Proceedings of IEEE/RSJ International Conference on Intelligent Robots and Systems}, vol.~56, no.~6, pp.~10663-10668, 2020.

  \bibitem{orzechowski2018bsl}
    P. F. Orzechowski, A. Meyer, and M. Lauer, ``Tackling Occlusions and Limited Sensor Range with Set-based Safety Verification,'' {\it Proceedings of International Conference on Intelligent Transportation Systems}, pp.~1729-1736, 2018.

  \bibitem{hu2021medical}
     Y. Hu,  H. Su,  J. Fu,  H. R. Karimi,  G. Ferrigno,  E. D. Momi, and  A. Knoll, ``Nonlinear Model Predictive Control for Mobile Medical Robot Using Neural Optimization,'' {\it IEEE Transactions on Industrial Electronics}, vol.~68, no.~12, pp.~12636-12645, 2021.
     
  \bibitem{1}
  W. Chung,  S. Kim, M. Choi,  J. Choi, H. Kim, C. Moon, and J. Song, ``Safe Navigation of a Mobile Robot Considering Visibility of Environment,''
  {\it IEEE Transactions on Industrial Electronics}, vol.~56, no.~10, pp.~3941-3950, 2009.

  \bibitem{2}
  D. Portugal, P. Alvito, E. Christodoulou, G. Samaras , and J. Dias, ``A Study on the Deployment of a Service Robot in an Elderly Care Center,''
  {\it International Journal of Social Robotics}, vol.~11, no.~2, pp.~317-341, 2019.

  \bibitem{3}
  T. Kurosaka and M. Kaneko, ``Autonomous Mobile Robot Selecting Optimum Path with Safe Speed Control in Consideration of Blind Area of Vision Sensors,''
  {\it IEEJ Transactions on Electronics, Information and Systems}, vol.~4, no.~4, pp.~356-364, 2015.

  \bibitem{4}
  K. Akiyoshi, D. Chugo, S. Muramatsu, S. Yokota, and H. Hashimoto,
  ``Autonomous Mobile Robot Navigation Considering the Pedestrian Flow Intersections,''
  {\it Proceedings of IEEE/SICE International Symposium on System Integration}, pp.~428-433, 2020.

  \bibitem{5}
  J. Yuan, S. Zhang, Q. Sun, G. Liu, and J. Cai,
  ``Laser-Based Intersection-Aware Human Following With a Mobile Robot in Indoor Environments,''
  {\it IEEE Transactions on Systems, Man, and Cybernetics: Systems}, vol.~51, no.~1, pp.~354-369, 2021.

  \bibitem{6}
  J. Higgins and N. Bezzo,
  ``Negotiating Visibility for Safe Autonomous Navigation in Occluding and Uncertain Environments,''
  {\it IEEE Robotics and Automation Letters}, vol.~6, no.~3, pp.~4409-4416, 2021.

  \bibitem{7}
  M. Kobayashi and N. Motoi, ``Local Path Planning Method Considering Blind Spots Based on Cost Map for Wheeled Mobile Robot,''
  {\it IEEJ Transactions on Industry Applications},~vol.~141, no.~8, pp.~598-605, 2021.

  \bibitem{rgbd-1}
    T. Kim, S. Lim, G. Shin, G. Sim, and D. Yun, ``An Open-Source Low-Cost Mobile Robot System With an RGB-D Camera and Efficient Real-Time Navigation Algorithm,'' {\it IEEE Access}, vol.~10, pp.~127871-127881, 2022.

  \bibitem{rgbd-2}
  S. Song, H. Lim, S. Jung, and H. Myung, ``G2P-SLAM: Generalized RGB-D SLAM Framework for Mobile Robots in Low-Dynamic Environments,'' {\it IEEE Access}, vol.~10, pp.~21370-21383, 2022.

  \bibitem{rgbd-3}
    A. Durand-Petiteville, E. Le Flecher, V. Cadenat, T. Sentenac, and S. Vougioukas, ``Tree Detection With Low-Cost Three-Dimensional Sensors for Autonomous Navigation in Orchards,'' {\it IEEE Robotics and Automation Letters}, vol.~3, no.~4, pp.~3876-3883, 2018.
        
  \bibitem{LD-1}
    H. Tang, X. Niu, T. Zhang, L. Wang, and J. Liu, ``LE-VINS: A Robust Solid-State-LiDAR-Enhanced Visual-Inertial Navigation System for Low-Speed Robots,'' {\it IEEE Transactions on Instrumentation and Measurement}, vol.~72, pp.~1-13, 2023.

  \bibitem{LD-2}
  B. Zhou, D. Xie, S. Chen, H. Mo, C. Li, and Q. Li, ``Comparative Analysis of SLAM Algorithms for Mechanical LiDAR and Solid-State LiDAR,'' {\it IEEE Sensors Journal}, vol.~23, no.~5, pp.~5325-5338, 2023.

  \bibitem{LD-3}
  J. Yin, D. Luo, F. Yan, and Y. Zhuang, ``A Novel Lidar-Assisted Monocular Visual SLAM Framework for Mobile Robots in Outdoor Environments,'' {\it IEEE Transactions on Instrumentation and Measurement}, vol.~71, pp.~1-11, 2022.
      
  \bibitem{dwa}
  D. Fox, W. Burgard, and S. Thrun, ``The Dynamic Window Approach to Collision Avoidance,''
  {\it Proceedings of IEEE International Conference on Robotics $\&$ Automation Magazine}, vol.~4, pp.~23-33, 1997.

  \bibitem{drl}
    M. Kamezaki, R. Ong, and S. Sugano, ``Acquisition of Inducing Policy in Collaborative Robot Navigation Based on Multiagent Deep Reinforcement Learning,'' {\it IEEE Access}, vol.~11, pp.~23946-23955, 2023.
    

  \bibitem{nav2}
    S. Macenski, F. Martín, R. White, and J. G. Clavero, ``The Marathon 2: A Navigation System,'' {\it Proceedings of IEEE/RSJ International Conference on Intelligent Robots and Systems}, pp.~2718-2725, 2020.

\end{thebibliography}
\end{document}